\documentclass[10pt,twocolumn,letterpaper]{article}

\usepackage[pagenumbers]{cvpr} 

\usepackage{algorithm}
\usepackage{algpseudocode}
\usepackage{multirow}
\usepackage{wrapfig}

\usepackage{soul}

\usepackage{subcaption}
\usepackage{makecell}

\def\sp{$S$}

\let\svthefootnote\thefootnote
\newcommand\freefootnote[1]{%
  \let\thefootnote\relax%
  \footnotetext{#1}%
  \let\thefootnote\svthefootnote%
}

\definecolor{cvprblue}{rgb}{0.21,0.49,0.74}
\usepackage[pagebackref,breaklinks,colorlinks,allcolors=cvprblue]{hyperref}

\def\paperID{ } 
\def\confName{CVPR}
\def\confYear{2026}

\title{Bias In, Bias Out? Finding Unbiased Subnetworks in Vanilla Models}

\author{Ivan Luiz De Moura Matos\textsuperscript{1} \quad 
Abdel Djalil Sad Saoud\textsuperscript{1} \quad
Ekaterina Iakovleva\textsuperscript{1}\\
Vito Paolo Pastore\textsuperscript{2,3} \quad
Enzo Tartaglione\textsuperscript{1}\\[5pt]
{\textsuperscript{1}LTCI, Télécom Paris, Institut Polytechnique de Paris, France}\\
{
\textsuperscript{2}MaLGa-DIBRIS, University of Genoa, Italy}\\
{
\textsuperscript{3}AI for Good (AIGO), Istituto Italiano di Tecnologia, Italy}
}

\begin{document}
\maketitle

\begin{abstract}
The issue of algorithmic biases in deep learning has led to the development of various debiasing techniques, many of which perform complex training procedures or dataset manipulation.
However, an intriguing question arises: is it possible to extract fair and bias-agnostic subnetworks from standard vanilla-trained models without relying on additional data, such as unbiased training set? 
In this work, we introduce Bias-Invariant Subnetwork Extraction (BISE), a learning strategy that identifies and isolates ``bias-free'' subnetworks that already exist within conventionally trained models, without retraining or finetuning the original parameters.
Our approach demonstrates that such subnetworks can be extracted via pruning and can operate without modification, effectively relying less on biased features and maintaining robust performance. Our findings contribute towards efficient bias mitigation through structural adaptation of pre-trained neural networks via parameter removal, as opposed to costly strategies that are either data-centric or involve (re)training all model parameters. 
Extensive experiments on common benchmarks show the advantages of our approach in terms of the performance and computational efficiency of the resulting debiased model.
The code is available at:\\ \url{https://github.com/ivanluizmatos/BISE}.
\freefootnote{This work has been accepted for publication at the \textit{IEEE/CVF Conference on Computer Vision and Pattern Recognition (CVPR) 2026}.}
\end{abstract}
\section{Introduction}
\label{sec:introduction}

Deep learning has recently revolutionized machine learning by automating feature extraction while surpassing human performance in tasks ranging from image recognition~\cite{he2016deep,xu2021artificial} to natural language processing~\cite{vaswani2017attention,devlin2019bert,brown2020language}. 
Its success relies on the ability to learn statistical patterns from the data. Although this data-driven approach enables flexibility, it can also introduce vulnerabilities, particularly when training data contains unintended shortcuts~\cite{geirhos2020shortcut,aru2023mind} or artifacts~\cite{meister2023gender}. In this case, instead of learning meaningful causal relationships, the model may learn shortcuts reflecting \textit{spurious correlations} present in the training set but possibly absent in the data distribution where the model is evaluated. 
This phenomenon, where the model learns to rely excessively on spurious attributes rather than on the features that are \textit{intrinsically} relevant to the task, is referred to as \textit{shortcut learning}~\cite{geirhos2020shortcut}.

\begin{figure}[!t]
    \centering
    \includegraphics[width=0.9\columnwidth]{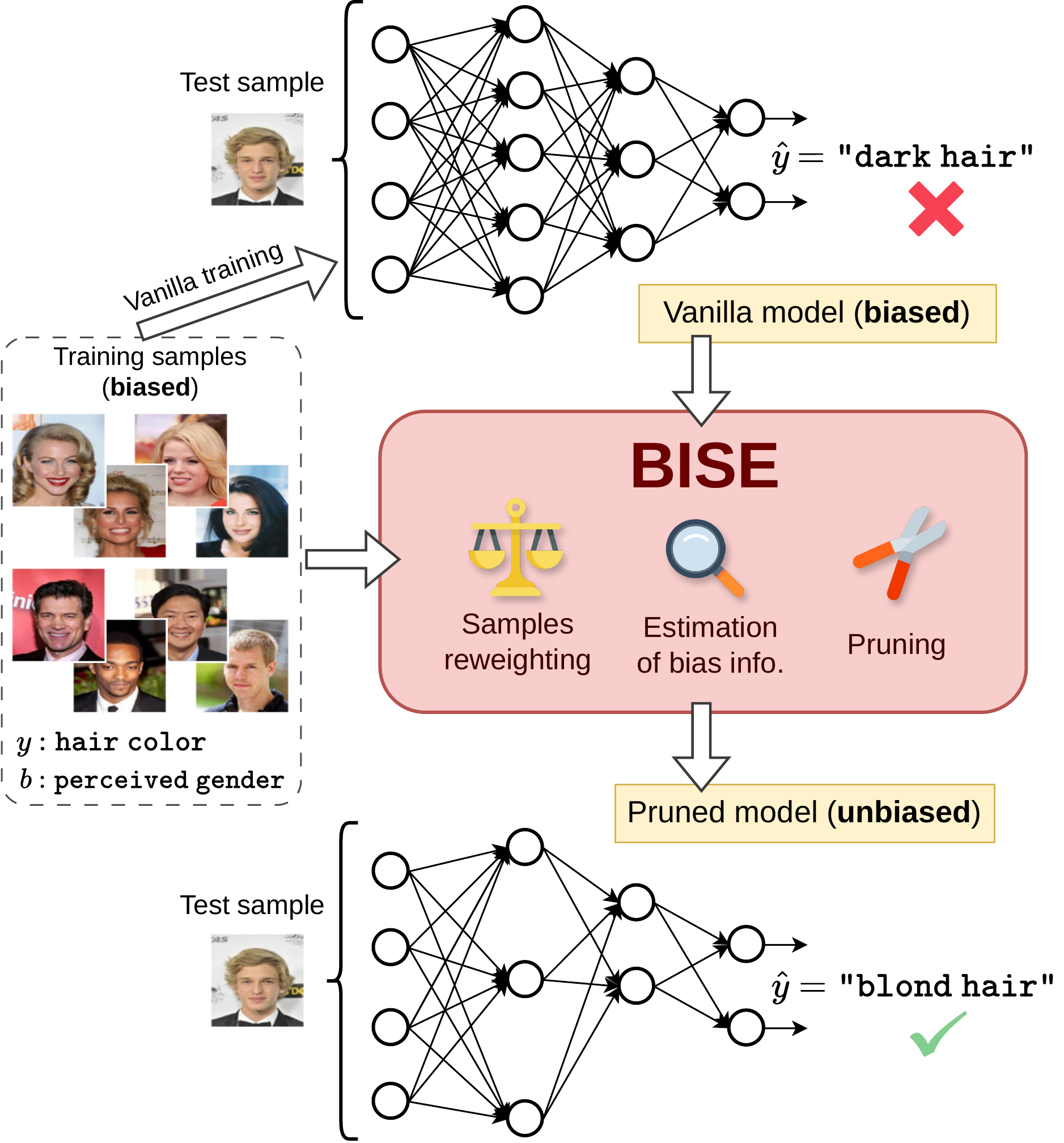}
    \caption{Overview of the proposed method. BISE aims to extract an unbiased subnetwork from the biased vanilla-trained network.}
  \label{fig:teaser}
\end{figure}

A critical manifestation of this limitation is \emph{algorithmic bias}, where models perpetuate or amplify unfair disparities~\cite{kordzadeh2022algorithmic}. Unlike societal biases (\emph{e.g.}, prejudiced human judgments), algorithmic biases arise implicitly from skewed data distributions or flawed optimization dynamics, generally without explicit discriminatory intent. For instance, a face recognition system might prioritize a certain demographic group
because it latches onto spurious correlations (like lighting conditions or corruptions).
In general, samples that verify spurious correlations (\emph{i.e.}, biases) are referred to as bias-\textit{aligned}, and they constitute the majority of the biased training set. The rest of the samples do not show these attributes and are denoted as bias-\textit{conflicting}. Such biases are particularly insidious because they can emerge even in applications where fairness is not the primary concern and where they are likely to be overlooked. Recently, the EU AI Act~\cite{act2024eu} intervened to regulate this problem, classifying high-risk AI systems (\emph{e.g.}, those used in hiring, law enforcement, or healthcare) as subject to strict transparency and fairness requirements, making algorithmic bias mitigation not just an ethical imperative but a legal one. 

In recent years, several debiasing methods have been proposed to mitigate model dependence on bias-related spurious correlations. 
Existing approaches are typically classified as \emph{data}-centric or \emph{model}-centric.
The first category,
which includes dataset rebalancing~\cite{zhao2023combating}, synthetic minority oversampling~\cite{liu2021just, pastore2025looking} and bias-conflicting sample augmentation~\cite{hwang2022selecmix, dataaug1}, aims to eliminate skewed representations at the input level. Model-centric methods, on the other hand, focus on the model itself, modifying, for example, the learning objective through adversarial debiasing~\cite{biasadv}, fairness constraints~\cite{li2022discover, sharma2024farfairnessconstraintshelp}, or bias-disentangled representation learning~\cite{end,barbano2023unbiased}. 
While often effective, it is required either (\textit{i}) the possibility to balance the data distribution via injection/exclusion of specific samples, typically not doable for bias-conflicting samples due to data scarcity, or (\textit{ii}) complete model retraining, computationally expensive and sometimes impractical for large-scale deployments. 
These methods typically treat bias as an external artifact that can be removed via additional training signals, rather than investigating whether standard models might already contain intrinsic bias-free representations that could be extracted through architectural interventions. Recent work has shown that, in principle, unbiased subnetworks can exist~\cite{nahon2024debiasing}, but they are obtained by leveraging unbiased variants of the original dataset, which is impractical in real scenarios.

In this work, we propose \textbf{B}ias-\textbf{I}nvariant \textbf{S}ubnetwork \textbf{E}xtraction (BISE), a paradigm shift that addresses these limitations by
extracting from vanilla-trained models the \textit{subnetworks} that are less biased than their original dense counterparts,
without requiring retraining or exploitation of bias-balanced training datasets (\cref{fig:teaser}).
BISE leverages this insight and shows that such subnetworks can be isolated through \textit{pruning}.
To extract more efficient debiased models, we resort to \textit{structured} pruning
(\textit{e.g.}, suppressing entire neurons or filters),
which enables network acceleration without requiring special hardware. 
Our experiments on popular benchmarks show how subnetworks extracted with BISE are competitive with the state of the art when used \textit{as is}, and are able to surpass it when further finetuned. 

Our contributions can be summarized as follows.
\begin{itemize}[leftmargin=0.5cm]
    \itemsep0em 
    \item We propose a method to identify and extract bias-robust subnetworks without requiring to retrain the model. This is performed by learning
    auxiliary variables that drive the extraction process (\cref{sec:method:learning_mask}).
    \item We design an objective function aligned with our goal. Specifically, we have both a balanced empirical loss term (Sec.~\ref{sec:loss_reweighting}) and a regularization term that reduces the amount of information related to the bias available to the task classifier (\cref{sec:mi_term}).
    \item Using popular benchmarks, we provide empirical evidence that unbiased subnetworks can be extracted based only on the original biased training set, without relying on additional bias-balanced datasets that are usually hard to obtain  (\cref{sec:experiments}), and how their accuracy can reach state-of-the-art when enhanced with finetuning. 
\end{itemize}

\section{Related work}
\label{sec:related}

\textbf{Debiasing.} Machine learning algorithms are usually developed with the assumption that the training set fully represents the target population, which might not be the case in reality. As a result, the model learns to rely on shortcuts \cite{geirhos2020shortcut} that stem from spurious correlations \cite{izmailov2022feature}, \emph{i.e.}, features that are predictive of the label (in the training set) but are not causally related to it. This negatively affects the generalization, group robustness, and fairness \cite{d2017conscientious,mehrabi2021survey,parraga2022debiasing} of the trained model. Debiasing paradigms can be roughly divided into three main categories: \textit{pre-processing}, \textit{in-processing}, and \textit{post-processing} \cite{d2017conscientious,mehrabi2021survey,wang2023biasing}. Pre-processing methods directly modify the data distribution in an attempt to mitigate representation bias, usually by means of data resampling \cite{li2019repair,agarwal2022does} or balancing heuristics \cite{kim2022learning}.
Instead of transforming the data, in-processing methods adapt the learning algorithm itself to account for biases during training, \textit{e.g.}, by using bias-aware optimization with balancing loss functions \cite{biascon,basu2024mitigating}, regularization \cite{bahng2020learning,sagawa2019distributionally} and contrastive learning \cite{barbano2023unbiased,koudounas2024contrastive,biascon,ma2021conditional,park2023training}, or by learning bias-invariant features with adversarial training \cite{zhang2024towards,kim2019learning,sarridis2024flac,passalis2018learning,ragonesi2021learning}. Finally, post-processing methods aim to mitigate biases within biased models
by finetuning \cite{gira2022debiasing}, distillation \cite{nam2020learning,blakeney2021simon}, or pruning \cite{nahon2024debiasing,meissner2022debiasing,marcinkevics2022debiasing}, for instance. Our method falls into the last category while successfully incorporating in-processing techniques, such as loss balancing and adversarial-like training.
In particular, our approach differs from debiasing via distillation, as the latter would generally require pre-defining the structure of a smaller, student, network, as in \cite{Delobelle2022FairDistillationMS, NEURIPS2022_79dc391a}.


\textbf{Pruning.} The recent success of deep learning models comes at the price of an increased parameter budget. In pursuit of reducing computational and memory burden, and to facilitate the deployment of models on devices with limited resources, researchers have considered various compression strategies, such as quantization, knowledge distillation, neural architecture search, and pruning \cite{cheng2024survey,frankle2018lottery}. There are three main types of pruning: \textit{unstructured} \cite{frankle2018lottery,sun2023simple,tanaka2020pruning,zhang2024towards}, where individual newtork parameters are removed, \textit{semi-structured} \cite{meng2020pruning,ma2020image}, where blocks of parameters are removed, and \textit{structured} \cite{nahon2024debiasing,zayed2024fairness,kong2024achieving,wang2019structured}, where the entire neurons, channels, filters or attention heads are removed. In this work, we focus on structured pruning,
as mentioned.
Among different pruning strategies, there are methods that rely on weight importance criteria to choose weights to prune, \textit{e.g.}, based on a score function, weight magnitude and norm, or loss gradient \cite{zhou2019deconstructing,sanh2020movement}. Another approach is to enforce sparsity through various regularization techniques \cite{wen2016learning,yuan2006model,gordon2018morphnet}. A popular view on pruning, which we also adopt, is to represent it with binary masks applied to the parameters of the neural network \cite{nahon2024debiasing,meissner2022debiasing,zhao2020masking}. These masks can be trained end-to-end using the initial loss function.


\textbf{Biases and compression.} Recently, several studies have investigated the effect of pruning on biases and fairness \cite{stoychev2022effect,ramesh2023comparative,jordao2021effect,iofinova2023bias}. It has been shown that, despite matching or being very close to uncompressed models in terms of overall performance, pruned networks tend to underperform on underrepresented or complex classes and cohorts \cite{hooker2019compressed,hooker2020characterising,paganini2020prune}, and have increased sensitivity to distribution shifts \cite{joseph2020going,liebenwein2021lost}. A significant scope of work has been dedicated to improving out-of-domain (OOD) generalization \cite{diffenderfer2021winning}, fairness \cite{meyer2022fair,zhang2024towards,zayed2024fairness,wu2022fairprune} and adversarial robustness \cite{liao2022achieving,sehwag2019towards} of pruned models, however, the main goal of these works is to preserve or only slightly improve the performance of the uncompressed model. In this work, we want our pruned subnetwork to behave differently from the significantly biased pre-trained network.

\section{Method}
\label{sec:method}

This section introduces BISE, our debiasing pruning method, which does not require additional training of the original model or access to unbiased training data. We start with the description of the setup (\cref{sec:method:setup}), followed by the design of our learnable pruning mask (\cref{sec:method:learning_mask}). Then we present the 
objective function, composed of a balanced empirical loss (\cref{sec:loss_reweighting}) and a regularization term that minimizes the impact of bias on the target task (\cref{sec:mi_term}). Finally, an overview of BISE is presented (\cref{sec:overview}).

\subsection{Setup}
\label{sec:method:setup}

We consider a common setup of \textit{supervised debiasing} for
classification~\cite{kim2019learning,end,bahng2020learning} where the dataset $\mathcal{D}$ consists of a \textit{biased} training set $\mathcal{D}_\text{train}$, and an \textit{unbiased} test set $\mathcal{D}_\text{test}$. Each input $x \in \mathcal{X}$ in $\mathcal{D}$ is associated with a pair of labels $(y,b)$, where $y \in \mathcal{Y}$ is the \textit{target label}, \emph{i.e.}, the attribute which the model is trained to predict, and $b \in \mathcal{B}$ is a \textit{bias} label, corresponding to an attribute of $x$ that is strongly correlated with $y$ in $\mathcal{D}_\text{train}$. 
Importantly, the labels $b$ and $y$ are not correlated in the unbiased $\mathcal{D}_\text{test}$, as there is generally no causal relationship between them. 
A network $f$
is therefore trained
to predict $y$; ideally, $f$ should not rely excessively on the attribute $b$.
In typical real-world applications, bias $b$ can represent a \textit{sensitive} attribute, such as gender, age, or ethnicity. For simplicity, we follow the literature~\cite{rubi,end,kim2022learning,nahon2024debiasing} and assume that there are $C$ target and bias classes ($C=\lvert\mathcal{Y}\rvert=\lvert \mathcal{B} \rvert$, where $\lvert\cdot\rvert$ denotes cardinality). 

Let $Y$, $\hat{Y}$, and $B$ denote the random variables associated with the ground-truth target $y$, predicted target $\hat y$, and bias label $b$, respectively.
For $~{(Y, \hat{Y}, B) \sim P_{\textrm{test}}}$, where $P_{\textrm{test}}$ denotes the data distribution of the unbiased test set $\mathcal{D}_\text{test}$, the network $f$ can be considered biased to feature $b$ if 
\begin{equation}
\label{eq:mutual_info_unequal}
    \mathcal{I}(\hat{Y}, B) \neq \mathcal{I}(Y, B),
\end{equation}
where $\mathcal{I}(\cdot, \cdot)$ denotes the \textit{mutual information} between two random variables~\cite{kim2019learning}.
For a biased model, \cref{eq:mutual_info_unequal} generally corresponds to $~{\mathcal{I}(\hat{Y}, B) \gg \mathcal{I}(Y, B)}$, which translates into an excessive reliance on the spurious feature $b$
when predicting $y$~\cite{tartaglione2022information,nam2020learning,barbano2024unsupervised, kim2019learning}. 

Inspired by~\cite{nahon2024debiasing}, we can opt to keep all parameters
in the biased pre-trained network unchanged and look for an unbiased subnetwork that \textit{already exists} within $f$. 
This provides us with the flexibility to account for biases in an already trained model (which were possibly unknown during vanilla-training) and ``correct'' them, without having to re-train the original, larger, network from scratch. 
Unlike~\cite{nahon2024debiasing}, we do not rely on an unbiased training dataset to debias and prune the pre-trained network $f$.

\subsection{Learning a debiasing pruning mask}
\label{sec:method:learning_mask}

We decompose our network as $f=\mathcal{C}\circ\mathcal{E}:\mathcal{X}\rightarrow \mathbb{R}^C$, where $~{\mathcal{E}=\mathcal{E}_\theta:\mathcal{X}\rightarrow\mathbb{R}^d}$ is an encoder, parameterised by $\theta \in \Theta$, and $\mathcal{C}$ is a classifier with latent representation of dimension $d$.
The output of $\mathcal{E}$ is considered a \textit{bottleneck}, and its latent representation of $x$ is denoted as $\hat{z}$.
Our goal is to \textit{learn} a mask $\mathcal{M}$ applied to the pre-trained and fixed parameters $\theta$, so that the representation $\hat{z} = \mathcal{E}_{\mathcal{M}(\theta)}(x)$ in the pruned model is informative of $y$ (\emph{i.e.}, enables accurate classification on $\mathcal{D}_\text{test}$)
and is less dependent on $b$.

As we propose to perform \textit{structured pruning},
we associate each structural component (\emph{e.g.}, a neuron or a filter) in the original encoder -- or rather the outputs of the following nonlinearities -- with a \textit{masking} parameter $m_i$. 
Let $h_i$ be the output of the $i$-th neuron/filter in the encoder of the original network $f$. Then, we modify each $h_i$ as:
\begin{equation}
\label{eq:gating}
    \hat{h}_i = h_i \cdot \mathbf{1}\{\hat{m}_i \geq 0.5\}, \;\; \text{with} \;\; \hat{m}_i = \sigma\Big(\frac{m_i}{\tau}\Big),
\end{equation}
where $\mathbf{1}\{\cdot\}$ denotes the indicator function, 
$\sigma(\cdot)$
is the usual \textit{sigmoid}, and $\tau > 0$. This gating mechanism is used to enforce $m_i$ to be further from zero, thus being more confident in the choice of the subnetwork. Here, $m_i$ is a trainable parameter initialized at zero, and $\tau$ is a \textit{temperature} that we anneal to zero as training progresses.
The masking parameter $m_i$ indicates whether the corresponding neuron/filter of the original network should be pruned ($m_i < 0$) or preserved ($m_i \geq 0$). 
Since our goal is to find an unbiased subnetwork that performs well in the unbiased test set $\mathcal{D}_\text{test}$ \textit{without} modifying any parameter of the original network $f$, we want to train the auxiliary parameters $\{m_i\}$ in an end-to-end manner.
In this case, the derivative of the step function $~{u(t) = \mathbf{1}\{t \geq 0\}}$ is zero for every $t \neq 0$ and is undefined at $t=0$; hence, to properly update $m_i$ by gradient descent, we adopt a
\textit{straight-through estimator}~\cite{bengio2013estimating}. 
We learn these masks by minimizing the following composite loss:
\begin{equation}
\label{eq:objective_function_general}
    J(\hat{y}, y, \hat{b}, b) = \mathcal{L}_{\text{CE}}(\hat{y},y) + \gamma \mathcal{I}(\hat{y},b).
\end{equation}
Here, $\mathcal{L}_{\text{CE}}(\hat{y},y)$ is a cross-entropy loss that accounts for the performance of the pruned model on the target task (\emph{i.e.}, classification of $y$), while $\mathcal{I}(\hat{y},b)$ measures the amount of private information related to $b$ that can be extracted from the bottleneck layer, and $\gamma > 0$ is a hyperparameter that controls the importance given to the second term. 
For a model that is accurate in predicting $y$, $~{\mathcal{I}(\hat{y},b) \approx \mathcal{I}(y,b)}$ is \textit{high} in $\mathcal{D}_\text{train}$, given the prevalence of bias-aligned samples (\textit{i.e.}, $y$ and $b$ are highly correlated, by construction). Hence, in \cref{eq:objective_function_general}, we
propose to compute 
$\mathcal{I}(\hat{y},b)$ only across the bias-conflicting samples, as in~\cite{nahon2023mining}.

We highlight that during the debiasing process, the parameters of the original network $f$ are kept intact, and only the mask parameters $\{m_i\}$ are trained. Importantly, we also do not require finetuning the pruned network after training the masks, although such a step can be \textit{optionally} performed to further improve the subnetwork performance. The extracted subnetwork is expected to produce accurate predictions $\hat{y}$ while being more invariant to $b$.

\subsection{Reweighting the cross-entropy loss} 
\label{sec:loss_reweighting}

Although it is possible to use the cross-entropy loss function in \cref{eq:objective_function_general} as presented, optimizing it on biased data $\mathcal{D}_\text{train}$ could negate the intended debiasing effect~\cite{kimimproving,geirhos2020shortcut,tsirigotis2023group}
(due to the significant prevalence of bias-aligned samples),
\textit{i.e.}, potentially leading to a biased subnetwork that performs well in the distribution of $\mathcal{D}_\text{train}$, but poorly on an unbiased set.
To discourage the debiasing algorithm from choosing a subnetwork that relies too heavily on spurious correlations, we replace $\mathcal{L}_{\text{CE}}(\hat{y},y)$ with its balanced version $\mathcal{L}_r(\hat{y},y)$ where the contribution of the bias-conflicting samples is amplified,
by assigning to each group $(y,b)$ a weight inversely proportional to its size \cite{sagawa2019distributionally, pmlr-v119-sagawa20a}.
In the setup considered in \cite{nam2020learning}
(\textit{i.e.}, where
$|\mathcal{Y}|= |\mathcal{B}|=C$, and $~{\mathbb{P}(y=\mathsf{y})=1/C , \,\mathbb{P}(\text{bias-aligned}\,| \, y=\mathsf{y})=\rho}$, $\forall \mathsf{y} \in \mathcal{Y}$, in $\mathcal{D}_\text{train}$), we can write:
\begin{equation}
\label{eq:balanced_ce}
    \mathcal{L}_r(\hat{y},y) = \frac{1}{N} \sum_{j=1}^{N} \ell(\hat{y}_j, y_j)\cdot r_j \; , 
\end{equation}
with $r_j =\frac{1}{C \rho}$ if $j$-th sample is bias-aligned, or $r_j =\frac{C-1}{C (1-\rho)}$ otherwise,
where
$\ell(\cdot,\cdot)$ is the per-sample cross-entropy loss,
$\rho$ denotes the proportion of bias-aligned samples in $\mathcal{D_\textrm{train}}$, $C$ is the number of target and private classes, and $N$ is the number of samples in the batch. Such balancing stems from two assumptions: (\textit{i}) the sum of weights over some set should be equal to its cardinality, and (\textit{ii}) all weights should be equal if the dataset is unbiased. The complete derivation of $r_j$ used in \cref{eq:balanced_ce}
can be found in the Supp. Mat.

\subsection{Minimizing the impact of bias on the target task}
\label{sec:mi_term}

With BISE, our goal is to identify and prune a subset of neurons that contribute disproportionately to the network's dependence on biased attributes, thereby promoting more balanced (debiased) representations. One core component of our approach, inspired by works on privacy preservation~\cite{tartaglione2022information,tartaglione2023disentangling},
consists in estimating how much information related to bias $b$ can be extracted and used by the pruned model while solving the target task (prediction of $y$). 

We consider a trained network $f=\mathcal{C}\,\circ\,\mathcal{E}$. To estimate the amount of bias information that can be leveraged by $\mathcal{C}$, we attach to the bottleneck (output of $\mathcal{E}$) an auxiliary classifier head 
$\mathcal{C}_{\text{aux}}$ of the same size as $\mathcal{C}$, and train it to predict the bias label $b$ from the latent embedding $\hat{z}$.
Let $\hat{B}$ be the random variable associated with the bias label $\hat{b}$ predicted with $\mathcal{C}_{\text{aux}}$.
If $\mathcal{C}_{\text{aux}}$ is a perfect classifier, then the mutual information between the predicted and ground-truth values of the private attribute, $~{\mathcal{I}(\hat{B},B)}$, is an upper bound of the information related to the bias that the classifier $\mathcal{C}$ can use to predict $\hat{y}$. This can be expressed also through $~{\mathcal{I}(\hat{B},B) \geq \mathcal{I}(\hat{Y},B)}$. Minimizing $~{\mathcal{I}(\hat{B},B)}$ will make the information related to the bias progressively harder to extract with $\mathcal{C}$. 
To properly estimate the bias information available at the bottleneck, and to continue to satisfy the upper bounding conditions, it is necessary to optimize the ability of $\mathcal{C}_{\text{aux}}$ to predict $\hat{b}$ from $\hat{z}$. For that, we follow~\cite{tartaglione2022information} and train $\mathcal{C}_{\text{aux}}$ by minimizing the cross-entropy loss $\mathcal{L}_{\text{CE}}(\hat{b},b)$ on $\mathcal{D}_\textrm{train}$.
With this change, the updated version of the composite loss from \cref{eq:objective_function_general} is: 
\begin{equation}
\label{eq:objective_function}
    J(\hat{y}, y, \hat{b}, b) = \mathcal{L}_r(\hat{y},y) + \gamma \mathcal{I}(\hat{b},b).
\end{equation}

\subsection{Overview of the approach}
\label{sec:overview}

\begin{algorithm}[h]
\caption{BISE($\mathcal{E}$, $\mathcal{C}$, $\mathcal{D}_{\text{train}}$, $E$,
$\kappa$, $\upsilon$, $\tau_{\text{min}}$)}
\label{alg:bise}

\resizebox{\linewidth}{!}{
\begin{minipage}{\linewidth}
\begin{algorithmic}[1] 
\State $m_i \gets 0, \, \forall \, i\,$; \quad $\tau \gets 1\,$; \quad epoch $\gets$ 0
\State Inject the masks: $\hat{h}_i = h_i \cdot \mathbf{1}\{ \sigma(\frac{m_i}{\tau}) \geq 0.5 \}$\label{line:1}
\State Instantiate $\mathcal{C}_\textrm{aux}$ as a fully connected layer with $|\mathcal{B}|$ neurons,
and attach $\mathcal{C}_\textrm{aux}$ to the output of $\mathcal{E}$
\State Train $\mathcal{C}_\textrm{aux}$ to minimize $\mathcal{L}_\text{CE}(\hat{b}, b)$ on $\mathcal{D}_{\text{train}}$, for $E$ epochs\label{line:2}
    \Repeat
        \State epoch $\gets$ epoch + 1
        \State On $\mathcal{D}_{\text{train}}$, for each minibatch:
            \State\hspace{\algorithmicindent} Compute $J(\hat{y}, y, \hat{b}, b)$ and $\mathcal{L}_\text{CE}(\hat{b}, b)$
            \State\hspace{\algorithmicindent} Update $\{ m_i \}$ to minimize $J(\hat{y}, y, \hat{b}, b)$
            \label{line:3}
            \State\hspace{\algorithmicindent} Update $\mathcal{C}_\textrm{aux}$ to minimize $\mathcal{L}_\text{CE}(\hat{b}, b)$
            \label{line:4}
        \If{epoch \% $\upsilon$ == 0}
            \State $\tau \gets \tau \cdot \kappa$\label{line:5}
            \State Finetune $\mathcal{C}_\textrm{aux}$ to min. $\mathcal{L}_\text{CE}(\hat{b}, b)$\label{line:6},
            for $E$ epochs
        \EndIf
\Until{$\tau < \tau_{\text{min}}$}\label{line:7}
\State \Return mask $\mathbf{1}\{ \sigma\big(\frac{m_i}{\tau_{\text{min}}}\big) \geq 0.5 \} \; \forall i $.
\end{algorithmic}
\end{minipage}
}
\end{algorithm}

\begin{figure}[t]
    \centering
    \includegraphics[width=0.9\columnwidth, clip, trim=0cm 0.3cm 0cm 0cm]{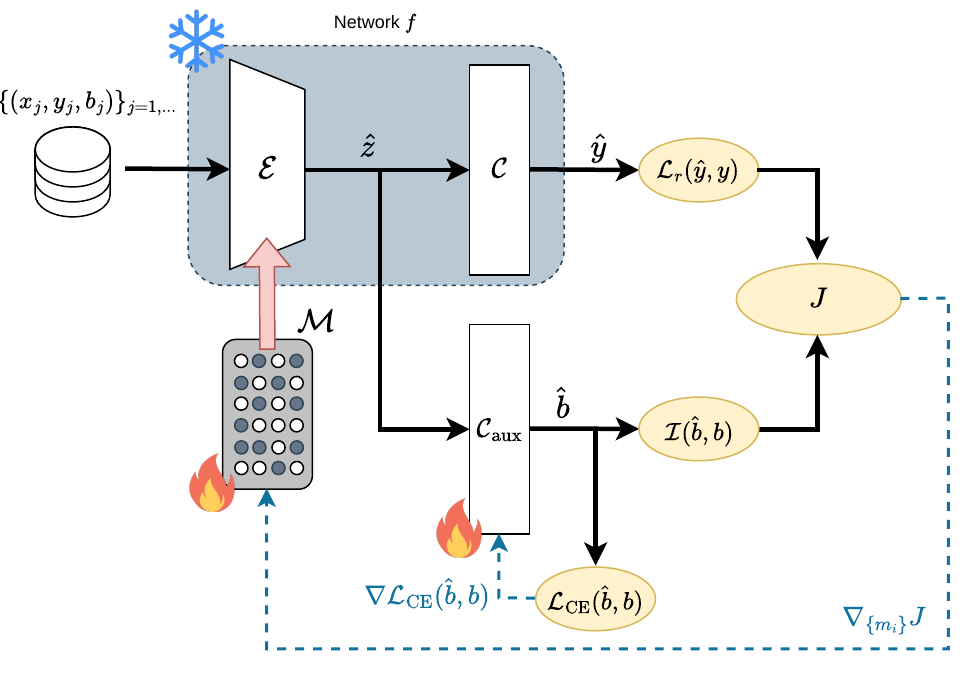}
    \caption{Illustration of BISE. Solid black arrows indicate forward propagation; dashed blue arrows indicate backward gradient propagation. During the training of the mask $\mathcal{M}$ and of $\mathcal{C}_\text{aux}$, the original model, $f = \mathcal{C}\circ\mathcal{E}$, is kept frozen.} 
    \label{fig:fig_pipeline}
\end{figure}

In Algorithm~\ref{alg:bise} we propose a summary of BISE. It requires as input the vanilla model (composed of the encoder $\mathcal{E}$ and the task classifier $\mathcal{C}$), the (biased) training set $\mathcal{D}_{\text{train}}$, and a set of hyper-parameters: the number $E$ of epochs required to train the auxiliary classifier $\mathcal{C}_\textrm{aux}$, a
factor $\kappa \in\, ]0,1[$ to anneal the temperature, the period $\upsilon$ (in epochs) for annealing the temperature, and the temperature $\tau_{\text{min}}$ at which point we stop BISE.
\Cref{fig:fig_pipeline} illustrates the approach.

After attaching the masking functions to the output of each neuron of the trained encoder (line~\ref{line:1}), we attach and train the auxiliary head to estimate usable information related to bias by $\mathcal{C}$ (line~\ref{line:2}). At this point, we begin learning the mask: we optimize both the masks (line~\ref{line:3}) and the auxiliary head $\mathcal{C}_\text{aux}$ (line~\ref{line:4}) for $\upsilon$ epochs, after which the temperature $\tau$ is decayed by a factor $\kappa$ (line~\ref{line:5}) and $\mathcal{C}_\text{aux}$ is refined (line~\ref{line:6}). The process is repeated, and ends when $\tau$ (initialized at 1) 
drops below $\tau_{\text{min}}$ (line~\ref{line:7}). The masks are then provided.
When an (unbiased) validation set $\mathcal{D}_\text{val}$ is available, we use it to select the best pruning mask, according to the validation accuracy. The results obtained without mask selection on the validation set are indicated as ``\textit{last}''.

\textbf{Finetuning.}
In the following, we show that the subnetwork extracted with BISE already achieves competitive results with respect to the state of the art. Although this is achievable without updating the weights of this subnetwork, the natural question arises then: how does the model behave if we continue to train this subnetwork on the same biased training set? To answer this, we also conducted finetuning experiments on the subnetworks extracted with BISE. When a validation set is available, we finetune the \textit{best} subnetwork found; otherwise, we finetune the subnetwork extracted in the last step of Algorithm~\ref{alg:bise}. The parameters of the subnetwork are updated by minimizing the reweighted cross-entropy loss $\mathcal{L}_r(\hat{y},y)$ on $\mathcal{D}_\text{train}$. Notably, finetuning does not affect the subnetwork size.

\section{Experiments}
\label{sec:experiments}

In this section, we provide the results obtained for BISE. After introducing the employed setup (\cref{sec:setup}), we will discuss how our method compares with respect to other debiasing approaches (\cref{sec:experiments:results}), and we analyze BISE's components with an ablation study (\cref{sec:experiments:ablation}). Our experiments are implemented with PyTorch 1.13.0., and performed on a server with a GPU NVIDIA RTX A4500 with 20 GB of VRAM. 
The results were averaged across three seeds.

\subsection{Experiments setup}
\label{sec:setup}

\textbf{Datasets.} For our experiments, we have selected five popular datasets, typically used for debiasing: (\textit{i}) \textit{BiasedMNIST}~\cite{bahng2020learning}, a synthetic dataset built on top of MNIST~\cite{726791}, where an injected background color acts as a bias; (\textit{ii}) \textit{Corrupted-CIFAR10}~\cite{hendrycks2019robustness}, a dataset built on top of CIFAR10~\cite{krizhevsky2009learning}, where the injected bias is a specific type of corruption; (\textit{iii}) \textit{CelebA}~\cite{liu2015deep}, where the perceived gender is a bias in our chosen classification task (prediction of the hair color); (\textit{iv}) \textit{Multi-Color MNIST}~\cite{li2022discover}, a benchmark for multiple biases handling, where the background of each image receives \textit{two} colors (one in the left-side, and one in the right); (\textit{v}) additionally, to evaluate BISE on a text classification setting, we chose \textit{CivilComments}~\cite{borkan2019nuanced, koh2021wilds}, where the task is to predict if sentences are toxic, while the bias is the presence of any ``sensitive'' element (we use the \textit{coarse} version of the dataset, as in~\cite{izmailov2022feature, idrissi2022simple, tiwari2024using}).
A more detailed description of the datasets 
can be found in the Supp. Mat.

\textbf{Architectures.} For all our experiments, we have used the same architectures as 
commonly considered in the literature: a convolutional neural network as described in \cite{bahng2020learning} for BiasedMNIST, a ResNet-18 pre-trained on ImageNet-1K from PyTorch~\cite{NEURIPS2019_bdbca288}
for both CelebA and Corrupted-CIFAR10~\cite{liu2015deep,hendrycks2019robustness}, and an MLP as in \cite{nam2020learning} for Multi-Color MNIST.
For CivilComments, we employ a BERT model pre-trained on Book Corpus and English Wikipedia~\cite{devlin2019bert}.
The details of these architectures and the related learning policies for the vanilla models are provided in the Supp.~Mat.
There, we also present the number of parameters $m_i$ that are trained during BISE, which represent a small fraction of the weights in the original dense model.

\textbf{Debiasing settings.} For learning the masks, we use SGD with learning rate $10^{-2}$, momentum $0.9$, and weight decay $10^{-4}$; $\tau$ is scaled by ${\kappa=0.5}$ every ${\upsilon=10}$ epochs, and the algorithm stops when $~{\tau < \tau_{\text{min}}=10^{-3}}$. The auxiliary classifier $\mathcal{C}_\textrm{aux}$ is trained with SGD, with learning rate $0.1$, momentum $0.9$, weight decay $10^{-4}$; $E$ is set to $50$ epochs, and $\gamma=1$. 
For Multi-Color MNIST, as we are in the presence of multiple biases, we follow the reweighting proposed in \cite{nahon2023mining}.
For the finetuning of the subnetwork extracted with BISE, we use the same optimizer as for training the vanilla model. Details are provided in the Supp. Mat.

\subsection{Main results}
\label{sec:experiments:results}

In this section, we discuss the results obtained with our proposed method.
To evaluate our method, we focus on the \textit{task accuracy} (Acc.), regarding the classification of $y$ on an \textit{unbiased} test set. We also report the \textit{sparsity} (\sp) of the extracted subnetworks, measured as the percentage of 
parameters
pruned from the original dense model,
as well as the \textit{computational complexity} ($\mathcal{O}$) at inference time,
measured in FLOPs.
Our primary objective is to be able to extract subnetworks that achieve higher task accuracy on an unbiased (test) set, compared to their vanilla dense counterparts.
The results are reported for BiasedMNIST (\cref{tab:table_biased_mnist}), Corrupted-CIFAR10 (\cref{tab:table_cifar10c}), CelebA (\cref{tab:table_celeba}), Multi-Color MNIST
(\cref{tab:table_multicolor_mnist}), and CivilComments (\cref{tab:table_civilcomments}).
The sparsity of the original dense model and of the subnetworks extracted by BISE at the best and last epochs are denoted by $S_\text{orig}$, $S_\text{BISE}$, and $S_\text{BISE}^\text{last}$, respectively. The corresponding complexities are $\mathcal{O}_\text{orig}$, $\mathcal{O}_\text{BISE}$, and $\mathcal{O}_\text{BISE}^\text{last}$, respectively.

Tables~\ref{tab:table_biased_mnist} and~\ref{tab:biased_mnist_sparsity_flops} summarize the results on BiasedMNIST for different proportions $\rho$ of bias-aligned samples in the training set. For all values of $\rho$ considered, BISE extracts a subnetwork that showcases a better test performance than the corresponding vanilla model, indicating successful reduction of the bias -- even in the condition of particularly strong spurious correlations, as for $\rho=0.997$. Additionally, we observe a significant reduction in the model's complexity due to pruning, as indicated by the sparsity and number of FLOPs. It is worth noticing that the models obtained by all the other methods presented in \cref{tab:table_biased_mnist} have the same inference complexity as the vanilla model, since such methods are not removing network parameters.

\begin{table}[!h]
\centering
\caption{Results on BiasedMNIST for different $\rho$ values.}
\label{tab:table_biased_mnist}
\resizebox{0.75\columnwidth}{!}{
\begin{tabular}{lccc}
\toprule
\multirow{2}{*}{\textbf{Method}}          & \multicolumn{3}{c}{\textbf{Accuracy (\%)}} \\
                                 \cmidrule{2-4}
                                 & $\rho=0.99$ & $\rho=0.995$ & $\rho=0.997$ \\
\midrule
Vanilla                          & $88.9_{\pm 0.4}$         & $75.1_{\pm 4.2}$         & $66.1_{\pm 1.7}$      \\
\midrule
Rubi \cite{rubi}                 & 93.6              & 43.0             & 90.4                   \\
EnD \cite{end}                   & 96.0             & 93.9         & 83.7                 \\
BCon+BBal \cite{biascon}         & \textbf{98.1}          & \textbf{97.7}          & \textbf{97.3}               \\
ReBias \cite{bahng2020learning}             & 88.4         & 75.4           & 65.8            \\
LearnedMixin \cite{clark2019don} & 88.3          & 78.2           & 50.2              \\
LfF \cite{nam2020learning}                   & 95.1         & 90.3           & 63.7            \\
SoftCon \cite{biascon}           & 95.2          & 93.1         & 88.6          \\
\midrule

BISE  & $96.1_{\pm 0.5}$         & $92.2_{\pm 1.9}$     & $90.8_{\pm 0.6}$        \\

BISE + finetuning    & $\bf 98.1_{\pm 0.1}$       & $\underline{96.3_{\pm 0.7}}$       & $\underline{95.9_{\pm 0.4}}$    \\

BISE (last)  & $95.7_{\pm 0.6}$        & $91.2_{\pm 1.9}$      & $88.3_{\pm 3.2}$    \\

\bottomrule
\end{tabular}
}%
\end{table}


\begin{table}[!h]
\centering
\caption{Sparsity (\sp) and computational cost (MFLOPs) of models, for the BiasedMNIST dataset.}
\label{tab:biased_mnist_sparsity_flops}
\resizebox{\columnwidth}{!}{
\begin{tabular}{
    l
    @{\hskip 8pt}
    cc
    @{\hskip 15pt}
    cc
    @{\hskip 15pt}
    cc
}
\toprule
\multirow{3}{*}{\textbf{Method}} 
& \multicolumn{6}{c}{\textbf{Proportion of bias-aligned samples in the training set ($\rho$)}} \\
\cmidrule(lr){2-7}
& \multicolumn{2}{c}{$\rho = 0.99$}
& \multicolumn{2}{c}{$\rho = 0.995$}
& \multicolumn{2}{c}{$\rho = 0.997$} \\
\cmidrule(lr){2-7}
& \sp\ (\%) $\uparrow$ & MFLOPs $\downarrow$
& \sp\ (\%) $\uparrow$ & MFLOPs $\downarrow$
& \sp\ (\%) $\uparrow$ & MFLOPs $\downarrow$ \\
\midrule
Vanilla & 0 & 415.4 & 0 & 415.4 & 0 & 415.4 \\
\midrule

BISE & $20.9_{\pm 4.4}$ & $328.3_{\pm 18.2}$ 
     & $29.9_{\pm 7.5}$ & $290.8_{\pm 31.0}$ 
     & $35.0_{\pm 1.5}$ & $269.6_{\pm 6.2}$ \\
    
BISE (last) & $18.9_{\pm 4.3}$ & $336.4_{\pm 17.7}$ 
     & $30.0_{\pm 6.8}$ & $290.7_{\pm 28.3}$ 
     & $33.8_{\pm 4.0}$ & $274.6_{\pm 16.7}$ \\

\bottomrule
\end{tabular}
}
\end{table}


On Corrupted-CIFAR10 (Tables \ref{tab:table_cifar10c} and \ref{tab:cifar10c_sparsity_flops}), BISE can find unbiased networks, better-performing than the original vanilla-trained model, for $\rho \in \{0.95, 0.98\}$, ranking as the best.
With subsequent finetuning, we obtain the highest accuracies across all values of $\rho$ considered.
\begin{table}[h]
    \centering
    \caption{Results on Corrupted-CIFAR10, for different values of $\rho$.
    }
    \label{tab:table_cifar10c}
\resizebox{\columnwidth}{!}{
    \begin{tabular}{lcccc}
    \toprule
    \multirow{2}{*}{\textbf{Method}}& \multicolumn{4}{c}{\textbf{Accuracy (\%)}} \\
    \cmidrule{2-5}
    & $\rho=0.995$ & $\rho=0.99$ & $\rho=0.98$ & $\rho=0.95$ \\
    \midrule
     
Vanilla                & $22.03_{\pm 0.39}$     & $26.89_{\pm 0.25}$     & $33.42_{\pm 0.65}$       & $47.18_{\pm 0.34}$                      \\
\midrule
EnD \cite{end}                       & 19.38         & 23.12         & 34.07       & 36.57    \\
HEX \cite{wang2019learning}          & 13.87        & 14.81          & 15.20    & 16.04     \\
ReBias \cite{bahng2020learning}                 & 22.27   & 25.72    & 31.66      & 43.43 \\
LfF \cite{nam2020learning}          & 28.57         & 33.07       & 39.91     & 50.27    \\
DFA \cite{dataaug1}           & \underline{29.95}       & \underline{36.49}     & 41.78      & 51.13    \\
\midrule

BISE (last)   & $21.94_{\pm 5.18}$   & $23.14_{\pm 11.42}$        & \underline{$42.80_{\pm 1.61}$}   & \underline{$55.38_{\pm 1.96}$}  \\

BISE (last) + finetuning   & $\bf{34.18_{\pm 2.14}}$     & $\bf{42.44_{\pm 0.36}}$      & $\bf{50.53_{\pm 1.13}}$   & $\bf{57.22_{\pm 1.81}}$  \\

    \bottomrule
    \end{tabular}
}%
    \end{table}


\begin{table}[!h]
\centering
\caption{Sparsity (\sp) and computational cost (MFLOPs) of models, for Corrupted-CIFAR10.}
\label{tab:cifar10c_sparsity_flops}
\resizebox{\columnwidth}{!}{
\renewcommand{\arraystretch}{1.25}
\setlength{\tabcolsep}{2pt} 
\begin{tabular}{
    l
    @{\hskip 8pt}
    cc
    @{\hskip 15pt}
    cc
    @{\hskip 15pt}
    cc
    @{\hskip 15pt}
    cc
}
\toprule
\multirow{3}{*}{\textbf{Method}} 
& \multicolumn{8}{c}{\textbf{Proportion of bias-aligned samples in the training set ($\rho$)}} \\
\cmidrule(lr){2-9}
& \multicolumn{2}{c}{$\rho = 0.995$}
& \multicolumn{2}{c}{$\rho = 0.99$}
& \multicolumn{2}{c}{$\rho = 0.98$}
& \multicolumn{2}{c}{$\rho = 0.95$} \\
\cmidrule(lr){2-9}
& \sp\ (\%) $\uparrow$ & MFLOPs $\downarrow$
& \sp\ (\%) $\uparrow$ & MFLOPs $\downarrow$
& \sp\ (\%) $\uparrow$ & MFLOPs $\downarrow$
& \sp\ (\%) $\uparrow$ & MFLOPs $\downarrow$ \\
\midrule
Vanilla & 0 & 37.1 & 0 & 37.1 & 0 & 37.1 & 0 & 37.1 \\
\midrule
\makecell[l]{BISE\\[-3pt](last)}
& $77.8_{\pm 2.3}$   & $15.5_{\pm 0.9}$      & $92.2_{\pm 1.0}$    & $15.7_{\pm 1.1}$      & $89.6_{\pm 0.6}$     & $19.2_{\pm 0.8}$   & $82.3_{\pm 0.6}$ & $22.5_{\pm 0.6}$  \\
\bottomrule
\end{tabular}
}
\end{table}


On CelebA (\cref{tab:table_celeba}), our method promotes again the debiasing of the vanilla network, despite not only the presence of spurious correlations between target and bias features, but also the significant target class imbalance known to affect the dataset \cite{Park_2024_CVPR}.
\begin{table}[!h]
\centering
\begin{minipage}{0.48\columnwidth}
\centering
\caption{Results on the CelebA dataset. 
}
\label{tab:table_celeba}
\resizebox{1\linewidth}{!}{
\begin{tabular}{lc}
\toprule
\textbf{Method} & \textbf{Acc. (\%)} $\uparrow$ \\
\midrule
    Vanilla & $76.5_{\pm 2.1}$ \\
\midrule
    EnD \cite{end} & 86.9 \\
    LNL \cite{kim2019learning} & 80.1 \\
    DI \cite{wang2020towards} & 90.9  \\
    BCon+BBal \cite{biascon} & \underline{91.4}  \\
    Group DRO \cite{sagawa2019distributionally} & 85.4  \\
    LfF \cite{nam2020learning} & 84.2  \\
    LWBC \cite{kim2022learning} & 85.5 \\
    CNC \cite{zhang2024correctncontrastcontrastiveapproachimproving} & 89.9 \\
\midrule

BISE &  $89.7_{\pm 0.8}$ \\

BISE + finetuning &  $\bf 91.8_{\pm 1.3}$ \\

BISE (last) &  $90.1_{\pm 0.9}$ \\

\bottomrule
\end{tabular}
}%
\end{minipage}
\hfill
\begin{minipage}{0.48\columnwidth}
\centering
\caption{Results on CivilComments. Results marked with $^\dagger$ are from~\cite{tiwari2024using}, $\ddagger$ is from~\cite{izmailov2022feature}, and $^\star$ are from~\cite{idrissi2022simple}.   
}
\label{tab:table_civilcomments}
\resizebox{1\linewidth}{!}{
\begin{tabular}{lccc}
\toprule
\textbf{Method} & \textbf{WGA (\%)} $\uparrow$ \\
\midrule
    Vanilla 
    & $59.6_{\pm 2.7}$  \\

\midrule
    Group DRO $\ddagger$ \cite{sagawa2019distributionally} & $\underline{80.4}$ \\
    JTT $^\star$\cite{liu2021just} & $67.8_{\pm 1.6}$ \\
    SimKD $^\dagger$ \cite{chen2022knowledge}  & $74.0_{\pm 2.25}$ \\
    DeTT $^\dagger$ \cite{lee2023debiased} & $75.0_{\pm 2.56}$ \\
    DeDiER $^\dagger$ \cite{tiwari2024using} & $78.3_{\pm 0.80}$ \\
    RWG $^\star$ \cite{idrissi2022simple} & $72.0_{\pm 1.9}$ \\
\midrule

BISE &  \underline{$80.4_{\pm 0.2}$}  \\

BISE + finetuning &  $\bf 81.0_{\pm 0.1}$  \\

BISE (last) &  $67.9_{\pm 1.8}$   \\

\bottomrule
\end{tabular}
}%
\end{minipage}
\end{table}

Regarding the sparsity, we have
$(S_\text{orig}, \; S_\text{BISE}, \; S_\text{BISE}^\text{last})=(0, \; 67.6_{\pm 0.8}, \; 67.7_{\pm 1.1}) \, \%$.
For the
complexity during inference, 
$(\mathcal{O}_\text{orig}, \; \mathcal{O}_\text{BISE}, \; \mathcal{O}_\text{BISE}^\text{last})= (1818.6, \; 821.5_{\pm 33.1}, \; 821.9_{\pm 32.3}) \, \text{MFLOPs}$.

In the case of multiple biases, here represented by the Multi-Color MNIST dataset (\cref{tab:table_multicolor_mnist}), we report, as Li \etal~\cite{li2022discover}, the test accuracy on four subgroups, identified as: (\textit{aligned, aligned}), (\textit{aligned, conflicting}), (\textit{conflicting, aligned}), and (\textit{conflicting, conflicting}), where the 1st and 2nd positions of the pair indicates if the samples in the group are bias-aligned \wrt the \textit{left}- and \textit{right}-side colors, respectively. We also show the \textit{unbiased accuracy}, \textit{i.e.}, the average across these groups.
BISE displays an improvement compared to the vanilla, according to the unbiased accuracy, while significantly improving the accuracy on the ``(\textit{conflicting, aligned})'' group. Interestingly, our approach is able to showcase a higher unbiased accuracy than FFW, which involves training on an unbiased set.
When finetuning the BISE-extracted subnetwork, we considerably improve the performance, in comparison to the vanilla model.
For the sparsity, we have $(S_\text{orig}, \; S_\text{BISE}^\text{last}) =(0, \; 17.1_{\pm 5.3}) \, \%$. Regarding the complexity during inference, $(\mathcal{O}_\text{orig}, \; \mathcal{O}_\text{BISE}^\text{last})= (256.2, \; 212.3_{\pm 13.5}) \, \text{kFLOPs}$.
\begin{table}[h]
\centering
\caption{Results on the Multi-Color MNIST dataset. (*) indicates that debiasing is performed by leveraging an unbiased dataset.}
\label{tab:table_multicolor_mnist}
\resizebox{\columnwidth}{!}{
\begin{tabular}{lcccccc}
\toprule
\multirow{2}{*}{\bf Method} 
& \multicolumn{5}{c}{\bf Accuracy (\%) $\uparrow$} \\
\cmidrule(lr){2-6}
& \small{(Alig., Alig.)} & \small{(Alig., Conf.)} & \small{(Conf., Alig.)} 
& \small{(Conf., Conf.)} & \small{Unbiased} \\
\midrule

Vanilla                & $\bf 100.0_{\pm 0.0}$        & $\underline{97.2_{\pm 0.8}}$        & $30.3_{\pm 1.4}$        & $5.2_{\pm 0.3}$        & $58.2_{\pm 0.6}$     \\
\midrule                                                                                              
LfF \cite{nam2020learning}               & $\underline{99.6}$        & 4.7         & $\bf 98.6$        & 5.1         & 52.0        \\
EIIL \cite{creager2021environment}             &{\bf 100.0}       & $\underline{97.2}$        & 70.8        & 10.9        & 69.7   \\
PGI  \cite{ahmed2020systematic}              & 98.6        & 82.6        & 26.6        & 9.5         & 54.3       \\
DebiAN  \cite{li2022discover}        & {\bf 100.0}       & 95.6        & 76.5        & 16.0        &\underline{72.0}    \\
VCBA \cite{nahon2023mining}             & {\bf 100.0}       & 90.9        & $\underline{77.5}$        & \underline{24.1}        & \textbf{73.1}    \\
FFW * \cite{nahon2024debiasing}             &   34.57     &   35.17      &    39.86     &   $\bf 35.85$      & 36.37  \\
\midrule

BISE  (last)  & $\bf 100.0_{\pm 0.0}$          & $88.8_{\pm 2.2}$          & $44.5_{\pm 2.3}$          & $7.7_{\pm 0.4}$          & $60.3_{\pm 1.0}$      \\

\makecell[l]{BISE (last)\\[-3pt]\hspace{2pt}+ finetuning}
& $\bf 100.0_{\pm 0.0}$  & $\bf 97.5_{\pm 0.7}$   & $71.1_{\pm 6.4}$   & $13.7_{\pm 1.8}$   & $70.6_{\pm 1.6}$    \\

\bottomrule
\end{tabular}%
}%
\end{table}

For CivilComments (\cref{tab:table_civilcomments}), we follow other works~\cite{izmailov2022feature, idrissi2022simple, tiwari2024using} and focus on the \textit{worst group accuracy} (WGA) metric.
Although WGA is not the same metric as employed for the other datasets, its improvement is associated with the reduction of spurious correlations, as discussed in~\cite{izmailov2022feature, idrissi2022simple}.
We observe that BISE is able to extract a subnetwork that achieves state-of-the-art WGA, notably when a validation set is leveraged to select the model.
For the sparsity,
$(S_\text{orig}, \; S_\text{BISE}, \; S_\text{BISE}^\text{last})=(0, \; 26.0_{\pm 5.4}, \; 45.7_{\pm 0.5}) \, \%$, measured as the percentage of pruned neurons in the feed-forward layers and pooler of BERT.

To summarize, although BISE is constrained to finding debiased models contained within biased dense networks without requiring finetuning (\emph{i.e.}, without the flexibility or granularity of training the original parameters of the vanilla model), it is generally competitive with the state of the art. Moreover, 
BISE leads to considerably smaller models and improved computational complexity at inference time, compared to the original networks
(as reported in Tab.~\ref{tab:biased_mnist_sparsity_flops} and Tab.~\ref{tab:cifar10c_sparsity_flops}, for instance), 
further strengthening its practical relevance. In addition, by subsequently finetuning the BISE-extracted subnetwork, we can further improve its performance, achieving even more competitive results.

\subsection{Ablation study}
\label{sec:experiments:ablation}

We present here some of the ablation studies that we have conducted for BISE. All the experiments are conducted on BiasedMNIST, with ${\rho=0.99}$, in the setup described in \cref{sec:setup}.
The sensitivity study on $E$, $\kappa$, $\upsilon$ and $\tau_{\text{min}}$ is provided in the Supp. Mat.


\textbf{Loss contribution.}
Here we ablate on the effect of the loss employed for BISE.
More specifically, Tab.~\ref{tab:table_ablation_loss} displays the influence of applying reweighting to the cross-entropy loss, and of enabling/disabling the mutual information term. 
As we observe, while the loss reweighting is determinant to achieve good performance in an unbiased scenario, the mutual information term enhances sparsity as it better regularizes the bottleneck, suggesting a reduction in the retention of bias-related features.


\begin{table}[h]
\centering
\caption{Ablation study for the components of the objective $J$. 
}
\label{tab:table_ablation_loss}
\resizebox{.45\textwidth}{!}{
\begin{tabular}{ccccc}
\toprule
Reweighting of $\mathcal{L}_\text{CE}(\hat{y},y)$ & $\mathcal{I}(\hat{b},b)$  &  {\bf Acc. (\%) $\uparrow$} & \sp\ (\%) $\uparrow$ & \textbf{MFLOPs} $\downarrow$\\
\midrule

      -      &       -      &  $91.9_{\pm 0.7}$ &  $5.8_{\pm 2.6}$ &  $390.9_{\pm 10.6}$  \\
$\checkmark$ &       -      &  $96.0_{\pm 0.4}$ &  $18.5_{\pm 6.5}$ &  $338.2_{\pm 27.1}$  \\
      -      & $\checkmark$ &  $91.1_{\pm 0.5}$ &  $18.7_{\pm 0.5}$ &  $337.3_{\pm 2.0}$  \\
$\checkmark$ & $\checkmark$ &  $96.1_{\pm 0.5}$ &  $20.9_{\pm 4.4}$ &  $328.3_{\pm 18.2}$  \\

\bottomrule
\end{tabular}
}%
\end{table}

\begin{figure}[h]
    \centering
    \begin{subfigure}[b]{0.49\columnwidth}
        \centering
        \includegraphics[width=\linewidth]
        {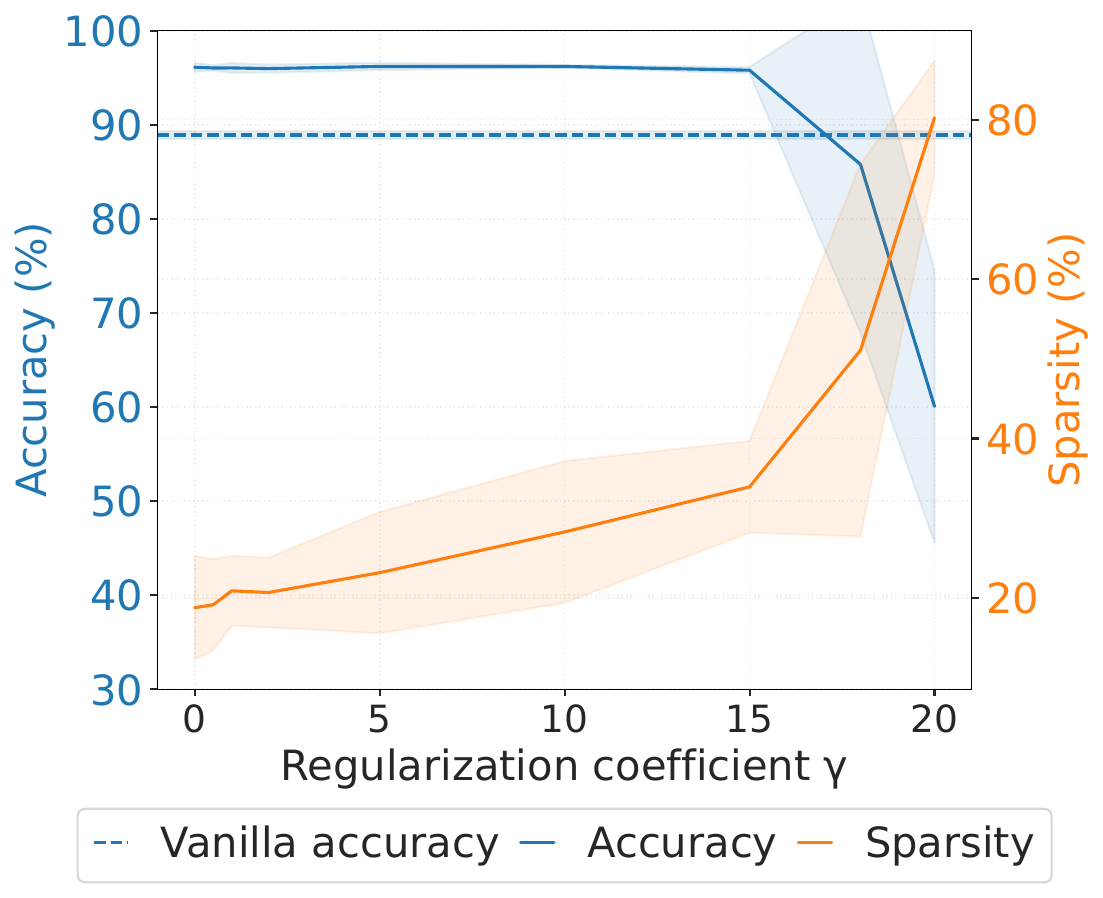}
        \caption{Effect of $\gamma$ on extracted model.}
        \label{fig:ablation_gamma}
    \end{subfigure}
    \hfill
    \begin{subfigure}[b]{0.495\columnwidth}
        \centering
        \includegraphics[width=\linewidth]
        {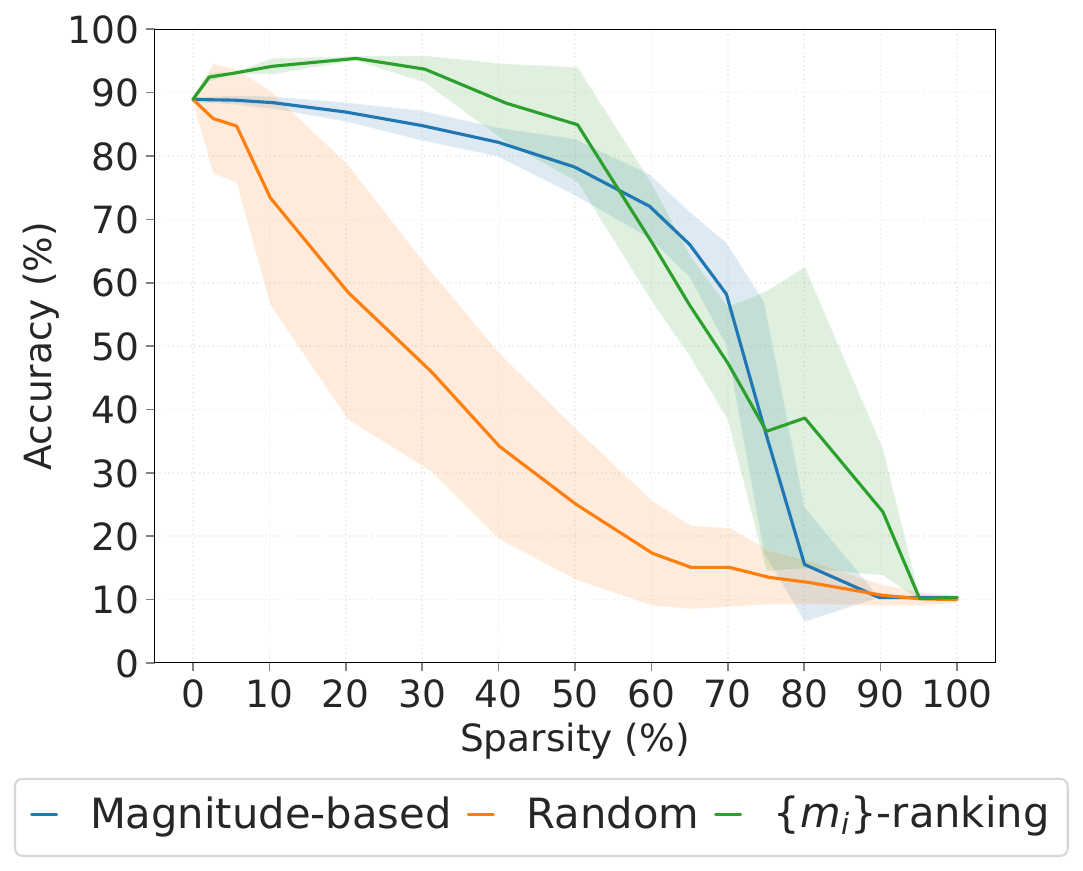}
        \caption{Comparison: pruning strategies.}
        \label{fig:pruning_comparison}
    \end{subfigure}
    \caption{Analysis of (a) coefficient $\gamma$ and (b) pruning strategies. Shaded areas indicate the interval of one standard deviation. 
    }
    \label{fig:combined_ablation_pruning}
\end{figure}

\textbf{Sensitivity to variations of $\bf \gamma$.} In \cref{fig:ablation_gamma}, we indicate
the influence of the regularization parameter $\gamma$ (see \cref{eq:objective_function}) on the
best
subnetwork that is found through the proposed method. In particular, $\gamma=0$ corresponds to not using the mutual information term, $\mathcal{I}(\hat{b},b)$, in the objective function $J$ minimized during the training of the mask; we observe that, in this case, we are still able to find an unbiased subnetwork, but with lower sparsity. Effectively, the results suggest that the mutual information term contributes to finding sparser models. We also note that, as $\gamma$ increases
indiscriminately, the performance of the extracted subnetwork drops. This may be explained by the fact that, for higher values of $\gamma$, $\mathcal{I}(\hat{b},b)$ may dominate $\mathcal{L}_r(\hat{y},y)$ in \cref{eq:objective_function_general}, thus strongly enforcing the obtention of a subnetwork that is invariant to $b$, regardless of its performance on the classification of $y$.

\textbf{Comparison with random and magnitude pruning.}
In Fig.~\ref{fig:pruning_comparison}, we compare the performance of the subnetworks extracted according to different (structured) pruning procedures: \textit{magnitude-based pruning} (where filters are ranked and removed according to the $L^2$-norm of their weights~\cite{10330640}), \textit{random pruning}, and BISE (here, filters are ranked and removed according to their corresponding $m_i$). We show results for different sparsity levels. It is important to highlight that we do not finetune the models considered.

For random pruning, accuracy rapidly drops as the sparsity increases, as expected, since parameters are removed without considering any importance measure. For magnitude pruning, the performance drops more slowly
and never surpasses the vanilla accuracy. In contrast, when pruning according to the auxiliary parameters $\{ m_i\}$ trained with BISE, we can extract a subnetwork that outperforms the vanilla model
in the unbiased test set (hence indicating successful bias mitigation),
for a certain range of sparsity levels -- specifically, around the sparsity level obtained when masking the network with $\mathbf{1}\{\sigma(\frac{m_i}{\tau}) \geq 0.5 \}$.
We note that, although magnitude pruning aims at identifying parameters whose removal is expected to least degrade model performance, this procedure does not result in an inherently unbiased subnetwork, in contrast to BISE.

\subsection{BISE
on unsupervised debiasing framework}

We now consider the \textit{absence} of bias labels $b$ during training, and follow a procedure to assign \textit{pseudo} bias labels $\tilde b$ to the samples (similarly to JTT~\cite{liu2021just}). We employ a secondary model (with the same architecture as the model $f$ to be debiased), which is vanilla-trained for a small number of epochs. This secondary model, $\tilde f$, is assumed to be strongly \textit{biased}, correctly classifying the \textit{bias-aligned} samples, while misclassifying \textit{bias-conflicting} ones
(as predictions $\tilde y$ would be strongly correlated with an underlying bias $b$).
This assumption follows from the observation (\textit{e.g.}, in~\cite{nam2020learning, nahon2024debiasing}) that bias-related features tend to be learned first. 

Each sample in $\mathcal{D}_\text{train}$ has its $\tilde b$ assigned as the corresponding prediction $\tilde y$.
We consider CelebA, and, for this specific study, we report the results across a set of three best seeds.
Model $\tilde f$ is trained for only \textit{one} epoch, as in \cite{liu2021just}.
The results are presented in \cref{tab:table_unsupervised}.
Despite the lack of true bias labels, BISE can still extract subnetworks that significantly outperform the vanilla on the unbiased test set. Comparing to \cref{tab:table_celeba}, we observe that we are competitive with other unsupervised debiasing approaches, \eg, LfF~\cite{nam2020learning}, LWBC~\cite{kim2022learning}.
To complement, in the Supp. Mat., we perform experiments in the unsupervised setting for BiasedMNIST.
\begin{table}[!h]
\centering
\caption{Results for unsupervised (\textit{Unsup.}) debiasing on CelebA.}
\label{tab:table_unsupervised}
\resizebox{0.9\columnwidth}{!}{
\begin{tabular}{lccc}
\toprule
\textbf{Method} & \textbf{Acc. (\%)} $\uparrow$ & \sp\ (\%) $\uparrow$ & \textbf{MFLOPs} $\downarrow$ \\
\midrule
    Vanilla & $76.5_{\pm 2.1}$ & 0  & 1818.6\\
\midrule

\textit{Unsup.} BISE 
& $87.5_{\pm 2.5}$
&  $67.3_{\pm 4.8}$ & $953.1_{\pm 218.4}$  \\

\textit{Unsup.} BISE (last) &  $85.1_{\pm 1.8}$ &  $67.7_{\pm 4.6}$ & $936.5_{\pm 190.1}$  \\


\bottomrule
\end{tabular}
}%
\end{table}


\subsection{Limitations}
\label{limit}
While our work advances the state of the art, the performance of our unbiased subnetworks is constrained by the compounding effects of multiple biases.
In particular, when biases interact, the pruned subnetworks \textit{as is} struggle to disentangle their contributions, though our subsequent finetuning effectively increases their robustness, at the price of 
increased computational burden. This suggests that further research is needed to assess our method in complex real-world scenarios, echoing findings in works like~\cite{li2022discover,nahon2023mining}, with retraining emerging as the only viable path to robustness. 
Scenarios
with a larger number of bias sources
(\eg, BiasedMNISTv2~\cite{Shrestha2022OccamNetsMD}) remain to be explored.
Furthermore, even though we empirically show that unbiased subnetworks exist in all the evaluated datasets with different
bias levels, the success of BISE is conditioned on the \textit{existence} of an unbiased subnetwork within the vanilla model. 
If such a substructure does not exist, 
BISE is not expected to extract a robust subnetwork.

\section{Conclusion
}
\label{sec:conclusion}

In this work, we have presented BISE, a method for debiasing neural networks through structured subnetwork extraction. We addressed the critical problem of algorithmic bias mitigation without requiring additional unbiased training set, or modifications to vanilla training procedures.

Unlike existing debiasing approaches that rely on model (re)training, our method identifies and extracts bias-resistant subnetworks within conventionally trained models through learning-based pruning and without altering the parameters of the vanilla model. 
This approach not only improves the performance of the model, but also slims it down, reducing both size and computation for its use.
We additionally showed that the performance of the subnetworks can be further improved via finetuning with a reweighted loss, reaching state-of-the-art performance on a collection of popular benchmark datasets in model debiasing.

Our findings suggest that bias mitigation can be achieved through purely architectural interventions, opening new directions for developing fairer models without expensive retraining or data curation. 

\section*{Acknowledgments}
This work was supported by the French National Research Agency (ANR) in the framework of the JCJC project “BANERA” under Grant ANR-24-CE23-4369, by Hi! PARIS and ANR/France 2030 program (ANR-23-IACL-0005).
Part of this work was funded by the European Union’s Horizon Europe Research and Innovation Programme under grant agreement No. 101120237 (ELIAS - European Lighthouse of AI for Sustainability).

{
    \small
    \bibliographystyle{ieeenat_fullname}
    \bibliography{main}
}

\clearpage
\maketitlesupplementary

\appendix

\numberwithin{equation}{section}
\numberwithin{figure}{section}
\numberwithin{table}{section}

\section{Discussion on the mutual information}

The mutual information terms $\mathcal{I}(\hat{B}, B)$ and $\mathcal{I}(\hat{Y}, B)$ can be expressed using entropy terms, $\mathcal{H}(\cdot)$, as follows:
\begin{equation}
    \begin{split}
    \mathcal{I}(\hat{B}, B) = \mathcal{H}(B) - \mathcal{H}(B|\hat{B}), \\
    \mathcal{I}(\hat{Y}, B) = \mathcal{H}(B) - \mathcal{H}(B|\hat{Y}).   
    \end{split}
\end{equation}
From these,
\begin{equation}
    \label{eq:mi_difference}
    \mathcal{I}(\hat{B}, B) - \mathcal{I}(\hat{Y}, B) = - \mathcal{H}(B|\hat{B}) + \mathcal{H}(B|\hat{Y}).
\end{equation}
In this work, we train a bias classification head $\mathcal{C}_{\text{aux}}$ on top of the encoder $\mathcal{E}$, as specified in 
\cref{sec:mi_term}.
If $\mathcal{C}_{\text{aux}}$ perfectly captures the bias encoded in the output of $\mathcal{E}$, \textit{i.e.}, bias classification accuracy $\approx 100\%$, then $\mathcal{H}(B|\hat{B}) \approx 0$. Since the conditional entropy is non-negative, from \cref{eq:mi_difference} we have:
\begin{equation}
    \mathcal{I}(\hat{B}, B) = \mathcal{I}(\hat{Y}, B) + \mathcal{H}(B|\hat{Y}) \geq \mathcal{I}(\hat{Y}, B).
\end{equation}
As a result, $\mathcal{I}(\hat{B}, B)$ remains an upper bound of $\mathcal{I}(\hat{Y}, B)$, as long as the encoder output contains enough information about the bias $B$ for the classifier $\mathcal{C}_{\text{aux}}$ to leverage. This also motivates periodic updates of the classifier $\mathcal{C}_{\text{aux}}$ to preserve the dependence between $\hat{B}$ and $B$.

Our setting might be viewed from another perspective. If we assume that our bias classifier $\mathcal{C}_{\text{aux}}$ is (nearly) perfect, 
which implies
$\mathcal{H}(B|\hat{B}) \approx 0$, then we can replace $\hat{B}$ with $B$ in our definition of a biased model in 
\cref{sec:method:setup}.
The model can then be considered biased if $~{\mathcal{I}(\hat{Y}, \hat{B}) \gg \mathcal{I}(Y, B)}$. In fact, if the mutual information between the predicted $\hat{Y}$ and $\hat{B}$ is large, while the latter is a perfect prediction of the true $B$,
it is suggested
that the classifier $\mathcal{C}$ effectively predicts $\hat{Y}$ from the bias, and minimizing $\mathcal{I}(\hat{Y}, \hat{B})$ would help us to reduce the reliance on bias $B$. Considering (under the assumption of perfect $\mathcal{C}_\text{aux}$) that $~{\mathcal{I}(\hat{B}, B) = \mathcal{I}(\hat{B}, \hat{B}) = \mathcal{H}(\hat{B})}$, we can derive the following statement:
\begin{equation}
\begin{split}
    \mathcal{I}(\hat{Y}, B) & \approx \mathcal{I}(\hat{Y}, \hat{B}) = \mathcal{H}(\hat{B}) - \mathcal{H}(\hat{B}|\hat{Y}) \\
    & = \mathcal{I}(\hat{B}, B) - \mathcal{H}(\hat{B}|\hat{Y}) \leq \mathcal{I}(\hat{B}, B).
\end{split}
\end{equation}
Thus, minimizing $\mathcal{I}(\hat{B}, B)$ is 
aligned with the goal of minimizing $\mathcal{I}(\hat{Y}, B)$.

Regarding theoretical guarantees, in 
\cref{eq:objective_function},
$\mathcal{I}(\hat B,B)$ is implicitly assumed to be an upper-bound to $\mathcal{I}(\hat Y,B)$. This condition is true when $\mathcal{H}(B | \hat B) = 0$, for which it is sufficient that $\mathcal{C}_\text{aux}$ is a perfect classifier ($\hat B \equiv B$), as discussed above; the last condition, however, cannot be always ensured.


\section{Reweighting of the cross-entropy loss}
\label{sec:appendix:details_reweighting}

We assume that a model that has been trained by traditional empirical risk minimization (ERM) is susceptible to bias when there is a significant prevalence of bias-aligned samples in the training data~\cite{kimimproving, sagawa2019distributionally, tsirigotis2023group}. 
We consider a dataset with $C$ target classes $y \in \mathcal{Y}$ and $C$ bias classes $b \in \mathcal{B}$, where each $y$ is strongly correlated with one $b$.
We denote by $b_y$ the bias class correlated with the target class $y$.
To account for bias during training, we modify the cross-entropy loss function $\mathcal{L}_\textrm{CE}(\hat{y},y)$
by \textit{upweighting} the bias-conflicting group.

Let $\rho$ be the proportion of bias-aligned samples in the training set $\mathcal{D}_\textrm{train}$. We assume that the bias-conflicting samples in $\mathcal{D}_\textrm{train}$ are uniformly distributed across the pairs $\{(y, b) : y \in \mathcal{Y}, b \in \mathcal{B}, b\neq b_y\}$ -- as is the case in the BiasedMNIST~\cite{bahng2020learning} and Corrupted-CIFAR10~\cite{hendrycks2019robustness} datasets. 
Inspired by reweighting based on the inverse frequency of the groups~\cite{sagawa2019distributionally, kimimproving},
we assign weight $r^\| = \frac{k^\|}{\rho}$ to bias-aligned samples, and $r^\perp = \frac{k^\perp}{1-\rho}$ for bias-conflicting samples, where $k^\|$ and $k^\perp$ are constant factors w.r.t. $\rho$. Let $r_j$ denote the weight assigned to the \textit{j}-th sample. To find $k^\|$ and $k^\perp$, we impose two constraints:

(\textit{i}) $\sum_{j=1}^{|\mathcal{D_\textrm{train}}|} r_j = |\mathcal{D_\textrm{train}}|$ (normalization),
which can be rewritten as
\begin{equation}
\label{eq:reweight_cond1}
\begin{array}{c}
r^\|\rho|\mathcal{D_\textrm{train}}| + r^\perp(1-\rho)|\mathcal{D_\textrm{train}}|=|\mathcal{D_\textrm{train}}| 
\\[0.5em]
\iff r^\|\rho + r^\perp(1-\rho)=1
\iff k^\| + r^\perp=1.
\end{array}
\end{equation}

(\textit{ii}) If the dataset is unbiased, all samples should receive the same weight, \emph{i.e.}: ${\rho = \frac{1}{C} \implies r^\| = r^\perp}$, from where we write
\begin{equation}
\label{eq:reweight_cond2}
\frac{k^\|}{1/C} = \frac{k^\perp}{1-1/C}
\iff k^\perp = (C-1)k^\|.
\end{equation}
From Equations~\eqref{eq:reweight_cond1}~and~\eqref{eq:reweight_cond2}, we obtain $k^\|=\frac{1}{C}$ and $k^\perp=\frac{C-1}{C}$. Hence: $r^\|=\frac{1}{C \, \rho}$ and $r^\perp=\frac{C-1}{C\, (1-\rho)}$. 

\textbf{Choice of weights for the datasets considered.} In BiasedMNIST and Corrupted-CIFAR10, we have $C=10$, therefore: $r^\|=\frac{1}{10 \, \rho}$ and $r^\perp=\frac{9}{10 \, (1-\rho)}$.

In the training set of CelebA~\cite{liu2015deep}, we observe an imbalance in terms of distribution of the target class (\texttt{BlondHair})~\cite{biascon, Park_2024_CVPR}.
In effect, most samples are labeled as dark-haired. For that reason, while reweighting the loss, we wish to take into consideration not only the spurious correlation between the gender and the hair color,
but also the aforementioned class imbalance. Let $c_\textrm{B}$ denote the proportion of blond-haired samples in the training set. Let $\rho_\textrm{B}$ represent the proportion of women inside the blond-haired group, and let $\rho_\textrm{D}$ be the proportion of women inside the dark-haired group. Following the idea of loss reweighting based on inverse-frequency of the groups~\cite{sagawa2019distributionally, kimimproving}, we denote by 
\begin{equation}
\begin{cases}
r^\textrm{MD} = \frac{k^\textrm{MD}}{(1-\rho_\textrm{D})\,(1-c_\textrm{B})} \,,\\[0.5em]
r^\textrm{WD} = \frac{k^\textrm{WD}}{\rho_\textrm{D}\,(1-c_\textrm{B})} \,,\\[0.5em]
r^\textrm{MB} = \frac{k^\textrm{MB}}{(1-\rho_\textrm{B})\,c_\textrm{B}} \,,\\[0.5em]
r^\textrm{WB} = \frac{k^\textrm{WB}}{\rho_\textrm{B}\,c_\textrm{B}}  \,,
\end{cases}
\label{eq:weights_celeba}
\end{equation}
the weights assigned for samples identified as dark-haired men, dark-haired women, blond-haired men, and blond-haired women, respectively, where $k^\textrm{MD}$, $k^\textrm{WD}$, $k^\textrm{MB}$ and $k^\textrm{WB}$ are constant factors with respect to $\rho_\textrm{D}$, $\rho_\textrm{B}$ and $c_\textrm{B}$.

Similarly to the reasoning that led to Equations \eqref{eq:reweight_cond1} and \eqref{eq:reweight_cond2}, we set the following constraints for CelebA:

(\textit{i}) $\; \sum_{j=1}^{|\mathcal{D_\textrm{train}}|} r_j = |\mathcal{D_\textrm{train}}|$, which is equivalent to
\begin{equation}
\begin{array}{c}
    r^\textrm{MD} (1-\rho_\textrm{D})\,(1-c_\textrm{B}) +
    r^\textrm{WD} \rho_\textrm{D}\,(1-c_\textrm{B}) + \\[0.5em]
    + r^\textrm{MB} (1-\rho_\textrm{B})\,c_\textrm{B} +
    r^\textrm{WB} \rho_\textrm{B}\,c_\textrm{B}
    = 1 \\[0.5em]
    \iff k^\textrm{MD} + k^\textrm{WD} + k^\textrm{MB} + k^\textrm{WB}  = 1.
\end{array}
\label{eq:cond1_celeba}
\end{equation}

(\textit{ii}) $\;  \rho_\textrm{D} = \frac{1}{2} \implies r^\textrm{MD} = r^\textrm{WD}$, which in turn can be rewritten (from \cref{eq:weights_celeba}) as
\begin{equation}
    \frac{k^\textrm{MD}}{(1-\frac{1}{2})\,(1-c_\textrm{B})} 
    = \frac{k^\textrm{WD}}{\frac{1}{2}\,(1-c_\textrm{B})}
    \iff k^\textrm{MD} = k^\textrm{WD}.
\label{eq:cond2_celeba}
\end{equation}

(\textit{iii}) $\;  \rho_\textrm{B} = \frac{1}{2} \implies r^\textrm{MB} = r^\textrm{WB}$, which (from \cref{eq:weights_celeba}) is equivalent to
\begin{equation}
    \frac{k^\textrm{MB}}{(1-\frac{1}{2})\,c_\textrm{B}}
    = \frac{k^\textrm{WB}}{\frac{1}{2}\,c_\textrm{B}}
    \iff k^\textrm{MB} = k^\textrm{WB}.
\label{eq:cond3_celeba}
\end{equation}

(\textit{iv}) $\;  c_\textrm{B} = \frac{1}{2} \implies
    r^\textrm{MD} (1-\rho_\textrm{D})\,(1-\frac{1}{2}) +
    r^\textrm{WD} \rho_\textrm{D}\,(1-\frac{1}{2}) =
    r^\textrm{MB} (1-\rho_\textrm{B})\,\frac{1}{2} +
    r^\textrm{WB} \rho_\textrm{B}\,\frac{1}{2}
$ (\textit{i.e.}, if the training set contains the same amount of dark-haired and blond-haired samples, then the total weight assigned to the dark-hair group should be the same as the total weight of the blond-haired group). The second member of the implication is equivalent to
\begin{equation}
\begin{array}{c}
    r^\textrm{MD} (1-\rho_\textrm{D}) +
    r^\textrm{WD} \rho_\textrm{D} =
    r^\textrm{MB} (1-\rho_\textrm{B}) +
    r^\textrm{WB} \rho_\textrm{B}
    \\[0.5em]
    \iff
    k^\textrm{MD} + k^\textrm{WD}
    = k^\textrm{MB} + k^\textrm{WB}.
\end{array}
\label{eq:cond4_celeba}
\end{equation}

Solving the system defined by Equations~\ref{eq:cond1_celeba},~\ref{eq:cond2_celeba},~\ref{eq:cond3_celeba} and~\ref{eq:cond4_celeba} leads to $~{k^\textrm{MD}=k^\textrm{WD}=k^\textrm{MB}=k^\textrm{WB}=\frac{1}{4}}$. Finally, by substituting these constants in~\cref{eq:weights_celeba}, we obtain
\begin{equation}
\begin{cases}
r^\textrm{MD} = \frac{1}{4\,(1-\rho_\textrm{D})\,(1-c_\textrm{B})} \,,\\[0.5em]
r^\textrm{WD} = \frac{1}{4\,\rho_\textrm{D}\,(1-c_\textrm{B})} \,,\\[0.5em]
r^\textrm{MB} = \frac{1}{4\,(1-\rho_\textrm{B})\,c_\textrm{B}} \,,\\[0.5em]
r^\textrm{WB} = \frac{1}{4\,\rho_\textrm{B}\,c_\textrm{B}}.
\end{cases}
\label{eq:weights_celeba_final}
\end{equation}

In Multi-Color MNIST~\cite{li2022discover}, each image is associated with two bias labels: the left color $b_\textrm{L} \in \mathcal{B}_\textrm{L}$, and the right color $b_\textrm{R} \in \mathcal{B}_\textrm{R}$, with $|\mathcal{B}_\textrm{L}|=|\mathcal{B}_\textrm{R}|=C$. We assume that each bias label is individually correlated with the target label as before, in the single-bias case. Let $\rho_\textrm{L}$ and $\rho_\textrm{R}$ denote the proportion of bias-aligned samples w.r.t. $b_\textrm{L}$ and $b_\textrm{R}$, respectively. We follow~\cite{nahon2024debiasing} and assign the weights
\begin{equation}
\begin{cases}
r^{\|,\|} = \frac{1}{\rho_\textrm{L}\,\rho_\textrm{R}} \,,\\[0.5em]
r^{\|,\perp} = \frac{1}{\rho_\textrm{L}\,(1-\rho_\textrm{R})}\,,\\[0.5em]
r^{\perp,\|} = \frac{1}{(1-\rho_\textrm{L})\,\rho_\textrm{R}}\,,\\[0.5em]
r^{\perp,\perp} = \frac{1}{(1-\rho_\textrm{L})\,(1-\rho_\textrm{R})}  \,,
\end{cases}
\label{eq:weights_multicolor_mnist}
\end{equation}
to the samples in the groups (\textit{aligned, aligned}), (\textit{aligned, conflicting}), (\textit{conflicting, aligned}), and (\textit{conflicting, conflicting}), respectively, where the first position of the pair indicates whether the group is bias-aligned w.r.t. the label $b_\textrm{L}$, and the second position indicates whether it is bias-aligned w.r.t. $b_\textrm{R}$.

In CivilComments, the data is divided into four groups, according to the pairs $(y,b)$, and the loss reweighting is analogous to the one applied to CelebA.

\section{Description of the datasets}

We present here an extended description of the datasets employed for the experiments.

\begin{figure}[h]
\includegraphics[width=.8\columnwidth]{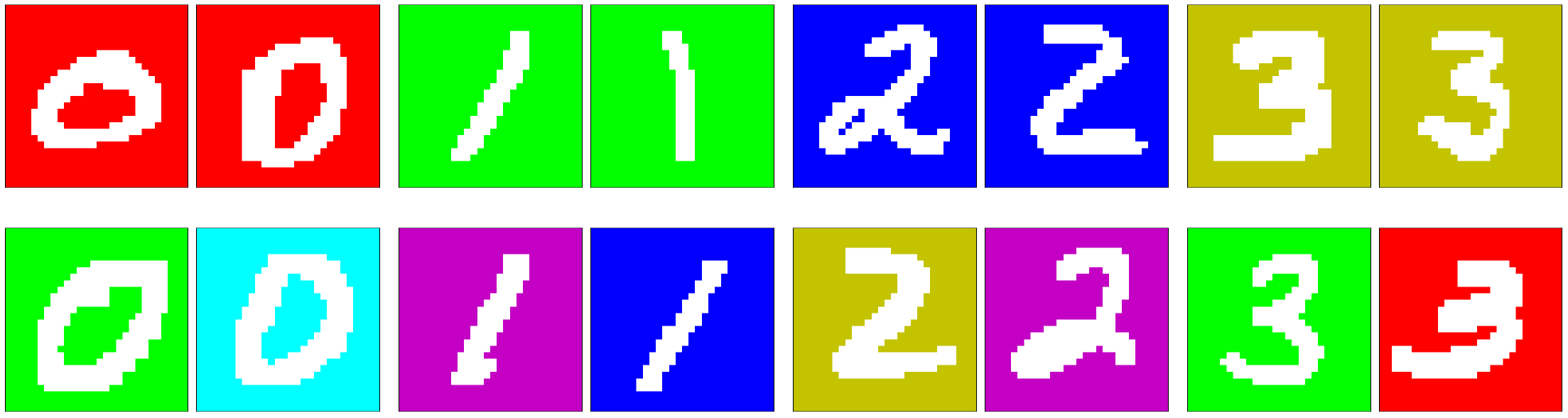}
\centering
\caption{Samples from the BiasedMNIST dataset~\cite{bahng2020learning}. In the first row, bias-aligned images; in the second row, bias-conflicting images.}
\label{fig_biased_mnist}
\end{figure}

\textbf{BiasedMNIST.}
Proposed by Bahng \textit{et al}.~\cite{bahng2020learning}, it consists of a synthetic dataset built upon the classic MNIST dataset~\cite{726791} by adding to the background of each image a color which is correlated with the target label in the following manner. Each digit $y \in \mathcal{Y} = \{0, 1, 2, \dots, 9 \}$ is associated with a unique \textit{predominant} background color $b_y \in \mathcal{B}$, where $|\mathcal{B}| = 10$, according to a ``one-to-one'' relation: for each image labeled as digit $y$, the background is colored with $b_y$ with probability $\rho \in [0,1]$, 
and with a random color $b \in \mathcal{B}\backslash\{b_y\}$ with probability $1 - \rho$.
Hence, the higher $\rho$, the stronger is the correlation between digits and colors. 
As commonly done in the literature for model debiasing, 
we build the training set $\mathcal{D}_\text{train}$ with large $\rho$,
while the unbiased test set $\mathcal{D}_\text{test}$ is built with $\rho=0.1$. Fig.~\ref{fig_biased_mnist} shows some
samples from the BiasedMNIST dataset.

\begin{figure}[h]
\includegraphics[width=.6\columnwidth]{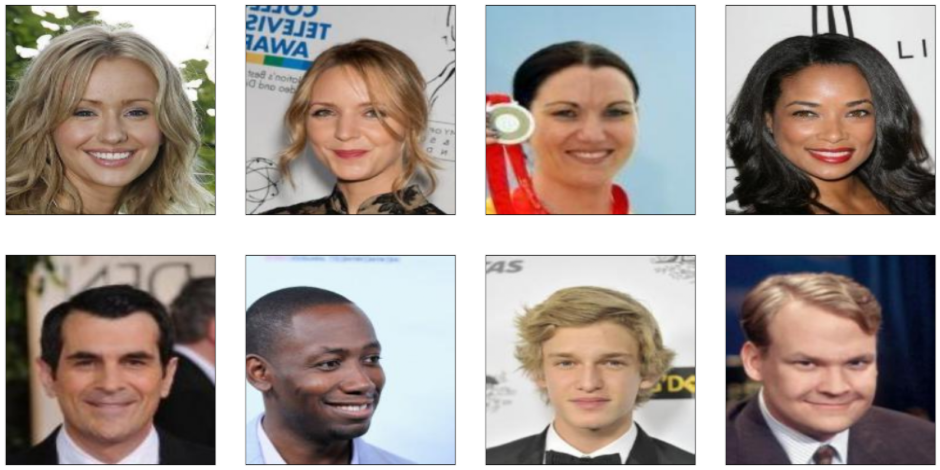}
\centering
\caption{Samples from the CelebA dataset~\cite{liu2015deep}.}
\label{fig_celeba}
\end{figure}

\textbf{CelebA.}
This real-world dataset, built by Liu \textit{et al}.~\cite{liu2015deep}, is composed of $202\,599$ face images, each described with 40 binary attributes. In CelebA, there exists a spurious relation between the perceived ``gender'' and the ``hair color'' attribute, with the majority of individuals with blond hair being labeled as women, and only a small percentage as men.
In our experiments, we consider the \texttt{BlondHair} attribute as our target, with the attribute \texttt{Male} as the bias, similarly to~\cite{sagawa2019distributionally, nam2020learning, tsirigotis2023group}.
In Fig.~\ref{fig_celeba}, we show samples corresponding to the four different combinations of the binary attributes \texttt{BlondHair} and \texttt{Male}.

\begin{figure}[h]
\includegraphics[width=.9\columnwidth]{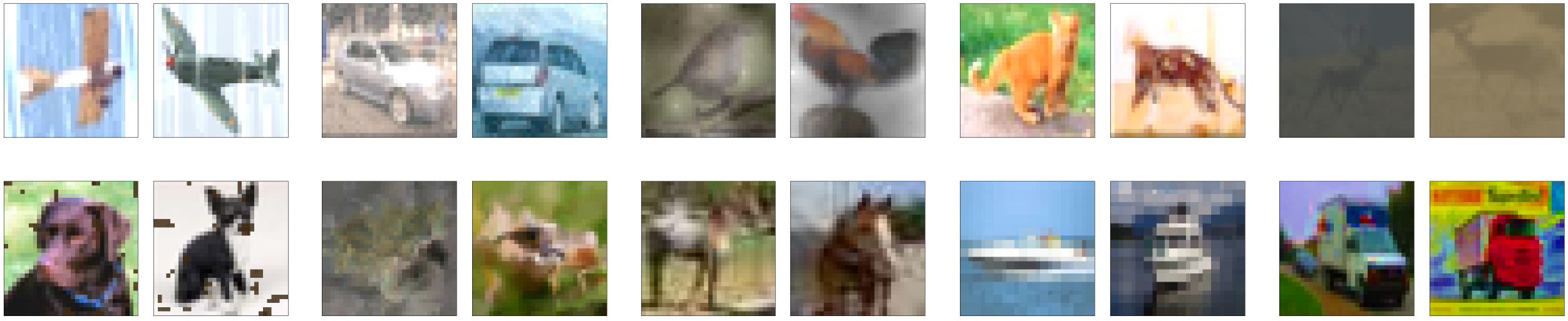}
\centering
\caption{Samples from the Corrupted-CIFAR10 dataset~\cite{hendrycks2019robustness}. We show two images belonging to each class.
}
\label{fig_cifar10c}
\end{figure}

\textbf{Corrupted-CIFAR10.} Developed by Hendrycks and Dietterich~\cite{hendrycks2019robustness}, this dataset is synthesized from the original CIFAR10 data~\cite{krizhevsky2009learning}, by applying to each image a \textit{corruption} (\eg, brightness, contrast, fog, etc.), and the type of corruption is correlated with the corresponding object class. In Fig.~\ref{fig_cifar10c}, we show samples from each class in the dataset.
As in BiasedMNIST, we denote by $\rho$ the proportion of bias-aligned samples in a set, and we consider $\rho \in \{0.95, 0.98, 0.99, 0.995\}$ for the training set, while the test data is unbiased ($\rho=0.1$). 
For each image, the target label $y$ is the object class, while the bias $b$ is the corruption that has been applied.

\begin{figure}[h]
\includegraphics[width=.8\columnwidth]{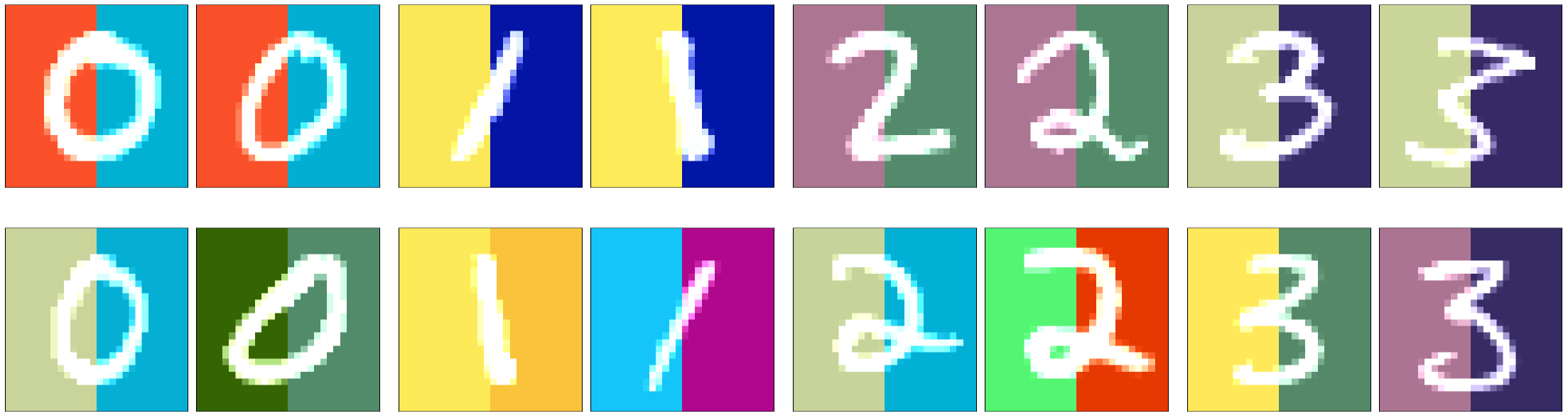}
\centering
\caption{Samples from the Multi-Color MNIST dataset~\cite{li2022discover}.  In the first row, images that are bias-aligned with respect to both background colors; in the second row, images that are bias-conflicting with respect to at least one of the colors.}
\label{fig_multicolor_mnist}
\end{figure}

\textbf{Multi-Color MNIST.}
When more than one bias is present, model debiasing becomes particularly challenging, as mitigating the reliance on one attribute may increase the dependency on another~\cite{kimimproving}.
To evaluate our method in such challenging cases, we employ \textit{Multi-Color MNIST}~\cite{li2022discover}, another dataset synthesized from MNIST. In this dataset, the background of each image consists of two colors: $b_\textrm{L}$ on the left side and $b_\textrm{R}$ on the right, which are correlated with the digit $y$ according to $\rho_\textrm{L}$ and $\rho_\textrm{R}$, respectively (see Fig.~\ref{fig_multicolor_mnist}). We use $(\rho_\textrm{L}, \rho_\textrm{R})=(0.99, 0.95)$ in the training set, and $(\rho_\textrm{L}, \rho_\textrm{R})=(0.1, 0.1)$ in the test set, as in~\cite{li2022discover, kimimproving}.
When applying BISE on Multi-Color MNIST, we employ \textit{two} auxiliary classifiers: one, $\mathcal{C}_\text{aux}^\text{L}$, to predict $b_\textrm{L}$, and another, $\mathcal{C}_\text{aux}^\text{R}$, to predict $b_\textrm{R}$.

\begin{figure}[h]
\includegraphics[width=.8\columnwidth]{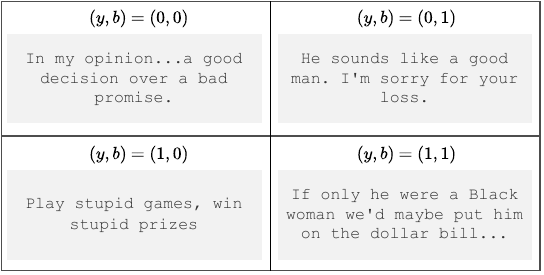}
\centering
\caption{Samples from the CivilComments dataset~\cite{borkan2019nuanced, koh2021wilds}. Label $y$ indicates whether a comment is toxic, while $b$ indicates if it mentions any sensitive attribute.}
\label{fig_civilcomments}
\end{figure}

\textbf{CivilComments}~\cite{borkan2019nuanced, koh2021wilds} is a text dataset composed of comments posted in online discussion forums. The associated task usually corresponds to classifying whether a comment is toxic. Besides having a binary label $y$ indicating toxicity, each comment is originally accompanied by binary labels $b_1,\dots,b_8$ indicating the presence of a mention to each of the sensitive attributes \textit{male}, \textit{female}, \textit{LGBT}, \textit{black}, \textit{white}, \textit{Christian}, \textit{Muslim}, \textit{other religion}~\cite{izmailov2022feature,idrissi2022simple}.
We follow Izmailov \etal~\cite{izmailov2022feature}, Idrissi \etal~\cite{idrissi2022simple} and Tiwari \etal~\cite{tiwari2024using}, and consider the \textit{coarse} version of the dataset, where the bias attributes are summarized into a single binary label $b$, which is equal to 1 if and only if at least one of the $b_1,\dots,b_8$ is 1.
Fig.~\ref{fig_civilcomments} shows some
samples from the CivilComments dataset.

\section{Architectures and training settings}

For BiasedMNIST, we employ the same architecture as in~\cite{bahng2020learning}: a fully convolutional network, composed of four layers of 7x7 kernels; each layer is followed by batch normalization and a ReLU activation. For training the vanilla model, the batch size is set to 100, and we use Stochastic Gradient Descent (SGD) with initial learning rate 0.1, momentum 0.9, and weight decay $10^{-4}$, as in~\cite{tartaglione2022information}; for the experiments with $\rho \in \{0.99, 0.995\}$, we train the vanilla model for 80 epochs, and the learning rate is divided by 10 at epochs 40 and 60.
For $\rho \in \{0.997, 0.999\}$, the vanilla model is trained for 100 epochs, and the learning rate is divided by 10 at epochs 80 and 90. 
During BISE, a masking parameter $m_i$ is assigned to each filter in the network.

For the experiments on CelebA and Corrupted-CIFAR10, we employ the ResNet18 architecture~\cite{he2016deep}, pretrained on ImageNet~\cite{ILSVRC15}. For CelebA, following~\cite{nam2020learning}, the vanilla model is trained for 50 epochs, using the Adam optimizer~\cite{KingmaB14} with learning rate $10^{-4}$, weight decay $10^{-4}$, and remaining hyperparameters as PyTorch's~\cite{NEURIPS2019_bdbca288} default values; the batch size is 256, and the training samples are augmented with random horizontal flip. For Corrupted-CIFAR10, the vanilla model is trained for 200 epochs, with batch size 256, and with the same optimizer, but with a learning rate initialized at $10^{-3}$ and reduced by cosine annealing, as in~\cite{barbano2023unbiased}; following~\cite{nam2020learning}, 32x32 random crop and random horizontal flip are applied to the training samples. 
When applying BISE, a masking parameter $m_i$ is assigned to each ReLU-activated output in ResNet18 residual building blocks.

For Multi-Color MNIST, we follow~\cite{li2022discover}: we employ an MLP with three hidden layers, each of them with 100 hidden units,
and, for the vanilla training,
we use the Adam optimizer with learning rate $10^{-3}$ and weight decay $10^{-4}$. We train the model for 100 epochs with batch size 256. During BISE, a mask parameter $m_i$ is assigned to each neuron in the hidden layers.

For CivilComments, the vanilla model is obtained by following Izmailov \etal~\cite{izmailov2022feature}: a BERT model (pre-trained on Book Corpus and English Wikipedia)~\cite{devlin2019bert} is trained on CivilComments for 10 epochs, using the AdamW optimizer~\cite{DBLP:conf/iclr/LoshchilovH19}, with learning rate initialized at $10^{-5}$ and linearly annealed, weight decay $10^{-4}$, and batch size 16. For applying BISE, we assign mask parameters $m_i$ to the neurons in the feedforward layers and the pooler layer in BERT.

\textbf{Finetuning settings.}
To finetune the BISE-extracted subnetwork, we employ the same optimizer that was used for training the vanilla model, with the original learning rate, except for BiasedMNIST and CivilComments, where the validation set is leveraged to select the learning rate: $10^{-3}$ for BiasedMNIST, Corrupted-CIFAR10 and Multi-Color MNIST, $10^{-4}$ for CelebA, and $10^{-8}$ for CivilComments. The finetuning is performed for 50 epochs, and, when a validation set is available, the best finetuned subnetwork is selected (according to the validation accuracy).

Experimental results reported for BISE (in the main paper and in the Supplementary Material) were obtained by averaging the results across three seeds.
When computing the sparsity of the BISE-extracted subnetworks for BiasedMNIST, CelebA, Corrupted-CIFAR10 and Multi-Color MNIST, we have employed the \textit{Simplify} library~\cite{bragagnolo2021simplify}, with batch normalization fusion enabled.


\section{On the number of parameters updated}
\Cref{tab:number_params_updated} shows the number of parameters $m_i$ that are trained during BISE, against the number of weights present in the corresponding dense model, for each dataset. The number of mask parameters $m_i$ is much smaller than the number of parameters in the network (since each $m_i$ is associated with one neuron/filter, not with an individual weight).
\begin{table}[h]
\centering
\caption{Number of weights \textit{vs.} of updated parameters $m_i$.
}
\resizebox{\columnwidth}{!}{
\begin{tabular}{lccccc}
\toprule

& \makecell[c]{\textbf{Biased}\\[-2pt]\textbf{MNIST}}

& \textbf{CelebA}

& \makecell[c]{\textbf{Corrupted-}\\[-2pt]\textbf{CIFAR10}}

& \makecell[c]{\textbf{Multi-Color}\\[-2pt]\textbf{MNIST}}

& \makecell[c]{\textbf{Civil
}\\[-2pt]\textbf{Comments}}
\\

\midrule
\# weights & $531\,210$ & $11\,177\,538$ & $11\,181\,642$ & $256\,510$ & $109\,361\,664$ \\
\# $m_i$ & $240 \; (0.05\%)$ & $3\,840 \; (0.03\%)$ & $3\,840 \; (0.03\%)$ & $300 \; (0.12\%)$ & $46\,848 \; (0.04\%)$ \\
\bottomrule
\end{tabular}
}
\label{tab:number_params_updated}
\end{table}


\section{Preliminary experiment}

As a preliminary experiment, we compare BISE with another method proposed for identifying unbiased subnetworks, FFW~\cite{nahon2024debiasing}. Although the main objective of FFW is simply to showcase the existence of unbiased subnetworks, which are identified leveraging an \textit{unbiased} training set (unrealistic setup), we believe it could be interesting to verify whether it could be applied in a case where the training set is \textit{biased}. 
Considering the BiasedMNIST dataset, when a biased training set is used and given a vanilla-trained model with test accuracy $88.9_{\pm 0.4} \, \%$, FFW can extract a subnetwork that displays a test accuracy of $80.6_{\pm 6.7} \, \%$, while the one extracted through BISE showcases an accuracy of
$96.1_{\pm 0.5} \, \%$
(see \cref{tab:table_biased_mnist}).
Hence, FFW fails to debias the vanilla network when a biased training set is leveraged.


\section{Results on BiasedMNIST for \texorpdfstring{$\bf \boldsymbol{\rho}=0.999$}{ρ=0.999}}

In Tab.~\ref{tab:table_biased_mnist_0.999}, we present the results obtained for our method on the BiasedMNIST dataset with $~{\rho=0.999}$. 

Firstly, we observe that, on the unbiased test set, the vanilla-trained model achieves accuracy corresponding to random guess (\textit{i.e.}, 10\%, as we have $C=10$ classes). This phenomenon suggests that the vanilla model essentially relies only on the bias-related features (the background color) to predict the digits, without effectively learning the core, relevant features (\textit{e.g.}, the digit shapes).

The subnetwork extracted with BISE achieves the accuracy of $15.7_{\pm 1.3} \, \%$. Although this is higher than the accuracy of the vanilla model, the proposed method could not isolate a well-performing subnetwork that would mostly rely on the features relevant to the task, without finetuning the remaining parameters. We also provide the results obtained with FFW~\cite{nahon2024debiasing}. The fact that FFW, a method that promotes debiasing by leveraging an \textit{unbiased} training set, also fails to identify a well-performing unbiased subnetwork suggests that such a subnetwork may not exist in the case of severe level of spurious correlations in the training set $\mathcal{D}_\textrm{train}$ (as it is the case here with $\rho=0.999$).
However, we show that, by further finetuning the BISE-extracted subnetwork, we can considerably improve its performance.
Effectively, as described in 
\cref{limit},
if an unbiased substructure does not exist within the vanilla model, then BISE is not expected to extract a robust subnetwork, which impacts the general performance of our approach.

\begin{table}[H]
\centering
\caption{Results on BiasedMNIST for $\rho=0.999$. (*) indicates that debiasing is performed by leveraging an unbiased dataset.}
\label{tab:table_biased_mnist_0.999}
\resizebox{0.9\columnwidth}{!}{
\begin{tabular}{lccc}
\toprule

\textbf{Method} & \textbf{Acc. (\%) $\uparrow$}  & \textbf{\sp\ (\%) $\uparrow$}  & \textbf{MFLOPs $\downarrow$} \\

\midrule
Vanilla            & $10.0_{\pm 0.1}$           & 0  & $415.4$     \\
\midrule

BISE  & $15.7_{\pm 1.3}$    & $85.6_{\pm 4.0}$  & $59.8_{\pm 16.7}$  \\

BISE + finetuning & $60.5_{\pm 9.4}$    & $85.6_{\pm 4.0}$  & $59.8_{\pm 16.7}$  \\

BISE (last)  & $14.7_{\pm 2.5}$         & $78.2_{\pm 2.9}$  & $90.5_{\pm 11.9}$  \\

\midrule

FFW * \cite{nahon2024debiasing}  & $19.9$         & --   & -- \\
                               
\bottomrule
\end{tabular}
}%
\end{table}


\section{Sensitivity analysis on \texorpdfstring{$E$}{E}, \texorpdfstring{$\kappa$}{κ}, \texorpdfstring{$\upsilon$}{υ} and \texorpdfstring{$\tau_{\text{min}}$}{τ\_min}}

In this section, we present a sensitivity study on the effect of the hyperparameters
$E$ (Tab.~\ref{tab:sensitivity_PH_epochs}), 
$\kappa$ (Tab.~\ref{tab:sensitivity_tau_update_factor}), 
$\upsilon$ (Tab.~\ref{tab:sensitivity_tau_update_period}) and 
$\tau_{\text{min}}$ (Tab.~\ref{tab:sensitivity_tau_min}) 
of the method proposed, on the BiasedMNIST dataset with $\rho=0.99$. As described in 
\cref{sec:setup},
in our experiments we used $E=50$ (the number of epochs for training/finetuning the auxiliary classifier $\mathcal{C}_\textrm{aux}$), $\kappa=0.5$ (the factor by which $\tau$ is updated), $\upsilon=10$ (the period of the $\tau$ updates) and $\tau_{\text{min}}=10^{-3}$ (the minimum value of $\tau$, which indicates when the algorithm should stop).

From the tables, we see that the proposed method demonstrates low sensitivity to variations in hyperparameters.
In particular, Tab.~\ref{tab:sensitivity_PH_epochs} indicates that, in the considered setup, there is no need to pre-train or finetune the auxiliary classifier $\mathcal{C}_\textrm{aux}$ for a large number of epochs. Additionally, Tables~\ref{tab:sensitivity_tau_update_factor} and~\ref{tab:sensitivity_tau_update_period} suggest that the rate with which the temperature $\tau$ is reduced (determined by the factor $\kappa$ and the period $\upsilon$) does not significantly impact the extracted subnetwork. Finally, Tab.~\ref{tab:sensitivity_tau_min} suggests that, after $\tau$ is reduced below $10^{-2}$, the algorithm has already 
found a certain unbiased subnetwork,
and executing the algorithm for longer does not lead to another, better-performing, subnetwork.

\begin{table}[h]
\centering
\caption{Effect of $E$ on the extracted subnetwork.}
\resizebox{\columnwidth}{!}{%
\begin{tabular}{lccccc}
\toprule
\multirow{2}{*}{\textbf{Metric}} & \multicolumn{5}{c}{$\boldsymbol{E}$} \\
\cmidrule{2-6}
& 1 & 5 & 10 & 20 & 50 \\
\midrule
\textbf{Acc. (\%)} $\uparrow$    & $95.9_{\pm 0.4}$ & $96.0_{\pm 0.5}$ & $96.0_{\pm 0.4}$ & $95.7_{\pm 0.5}$ & $96.1_{\pm 0.5}$ \\
\textbf{\sp\ (\%)} $\uparrow$     & $18.9_{\pm 7.3}$ & $17.6_{\pm 7.1}$ & $20.0_{\pm 6.1}$ & $17.2_{\pm 7.6}$ & $20.9_{\pm 4.4}$ \\

\textbf{MFLOPs} $\downarrow$     & $336.7_{\pm 30.4}$ & $342.1_{\pm 29.5}$ & $331.9_{\pm 25.2}$ & $343.7_{\pm 31.6}$ & $328.3_{\pm 18.2}$ \\

\bottomrule
\end{tabular}%
}
\label{tab:sensitivity_PH_epochs}
\end{table}


\begin{table}[h]
\centering
\caption{Effect of $\kappa$ on the extracted subnetwork.}
\resizebox{0.75\columnwidth}{!}{%
\begin{tabular}{lccc}
\toprule
\multirow{2}{*}{\textbf{Metric}} & \multicolumn{3}{c}{$\boldsymbol{\kappa}$} \\
\cmidrule{2-4}
& 0.1 & 0.5 & 0.8  \\
\midrule
\textbf{Acc. (\%)} $\uparrow$     & $96.0_{\pm 0.4}$ & $96.1_{\pm 0.5}$ & $96.2_{\pm 0.6}$ \\
\textbf{\sp\ (\%)} $\uparrow$     & $20.6_{\pm 3.9}$ & $20.9_{\pm 4.4}$ & $18.9_{\pm 7.6}$ \\

\textbf{MFLOPs} $\downarrow$     & $329.3_{\pm 16.3}$ & $328.3_{\pm 18.2}$ & $336.5_{\pm 31.4}$ \\

\bottomrule
\end{tabular}%
}
\label{tab:sensitivity_tau_update_factor}
\end{table}


\begin{table}[h]
\centering
\caption{Effect of $\upsilon$ on the extracted subnetwork.}
\resizebox{\columnwidth}{!}{%
\begin{tabular}{lccccc}
\toprule
\multirow{2}{*}{\textbf{Metric}} & \multicolumn{5}{c}{$\boldsymbol{\upsilon}$} \\
\cmidrule{2-6}
& 1 & 5 & 10 & 20 & 50  \\
\midrule
\textbf{Acc. (\%)} $\uparrow$     & $95.9_{\pm 0.5}$ & $95.8_{\pm 0.5}$ & $96.1_{\pm 0.5}$ & $96.0_{\pm 0.5}$ & $96.5_{\pm 0.3}$ \\

\textbf{\sp\ (\%)} $\uparrow$     & $22.4_{\pm 6.3}$ & $18.3_{\pm 6.2}$ & $20.9_{\pm 4.4}$ & $19.2_{\pm 6.3}$ & $19.8_{\pm 8.0}$ \\

\textbf{MFLOPs} $\downarrow$     & $322.2_{\pm 26.2}$ & $338.9_{\pm 25.7}$ & $328.3_{\pm 18.2}$ & $335.2_{\pm 26.2}$ & $333.0_{\pm 33.3}$ \\

\bottomrule
\end{tabular}%
}
\label{tab:sensitivity_tau_update_period}
\end{table}


\begin{table}[h]
\centering
\caption{Effect of $\tau_\textrm{min}$ on the extracted subnetwork.}
\resizebox{0.95\columnwidth}{!}{%
\begin{tabular}{lcccc}
\toprule
\multirow{2}{*}{\textbf{Metric}} & \multicolumn{4}{c}{$\boldsymbol{\tau_\textrm{min}}$} \\
\cmidrule{2-5}
& $10^{-1}$ & $10^{-2}$ & $10^{-3}$ & $10^{-4}$  \\
\midrule
\textbf{Acc. (\%)} $\uparrow$    & $95.9_{\pm 0.6}$ & $96.1_{\pm 0.5}$ & $96.1_{\pm 0.5}$ & $96.1_{\pm 0.5}$ \\

\textbf{\sp\ (\%)} $\uparrow$     & $20.4_{\pm 5.0}$ & $20.9_{\pm 4.4}$ & $20.9_{\pm 4.4}$ & $20.9_{\pm 4.4}$ \\

\textbf{MFLOPs} $\downarrow$    & $330.3_{\pm 20.7}$ & $328.3_{\pm 18.2}$ & $328.3_{\pm 18.2}$ & $328.3_{\pm 18.2}$ \\

\bottomrule
\end{tabular}%
}
\label{tab:sensitivity_tau_min}
\end{table}

\section{Updating batch normalization layers}
We have conducted an experiment where, besides learning the pruning mask, we are also updating the parameters in the batch normalization layers. It has been shown, indeed, that in some tasks like domain adaptation just updating these few parameters can be sufficient to have a significant gain in performance~\cite{olivi2025efficient}. In \cref{tab:table_bn_biased_mnist}, we show how learning the parameters in the  batchnorm layers 
-- specifically, updating the running statistics --
can prospectively boost the performance. We did not include this procedure in the main approach as it would require an increment in the number of learned parameters.
\Cref{tab:table_sparsity_flops_bn_ablation} displays the sparsity and complexity of the extracted subnetworks.

\begin{table}[!h]
\centering

\caption{Experiments on BiasedMNIST with trainable \textit{vs.} non-trainable batch normalization layers.
}
\label{tab:table_bn_biased_mnist}
\resizebox{\columnwidth}{!}{
\begin{tabular}{lccc}
\toprule
\multirow{2}{*}{\textbf{Method}}          & \multicolumn{3}{c}{\textbf{Accuracy (\%)}} \\
                                 \cmidrule{2-4}
                                 & $\rho=0.99$ & $\rho=0.995$ & $\rho=0.997$ \\

\midrule
Vanilla                          & $88.9_{\pm 0.4}$        & $75.1_{\pm 4.2}$         & $66.1_{\pm 1.7}$             \\

\midrule

BISE 
(\cref{tab:table_biased_mnist})
& $96.1_{\pm 0.5}$           & $92.2_{\pm 1.9}$        & $90.8_{\pm 0.6}$    \\

BISE + trainable BN (avg., std.)
& $97.3_{\pm 0.1}$    & $94.9_{\pm 1.1}$  & $94.0_{\pm 1.2}$ \\

BISE + trainable BN (avg., std., $\beta$, $\gamma$)
& $96.9_{\pm 0.4}$   & $92.2_{\pm 1.9}$ & $90.1_{\pm 3.7}$  \\

\bottomrule
\end{tabular}
}%
\end{table}


\begin{table}[!h]
\centering
\caption{Sparsity (\sp) and computational cost (MFLOPs) of models, for the BiasedMNIST dataset, when updating the batch normalization layers during BISE.}
\label{tab:table_sparsity_flops_bn_ablation}
\resizebox{\columnwidth}{!}{
\begin{tabular}{
    l
    @{\hskip 8pt}
    cc
    @{\hskip 15pt}
    cc
    @{\hskip 15pt}
    cc
}
\toprule
\multirow{3}{*}{\textbf{Method}} 
& \multicolumn{6}{c}{\textbf{Proportion of bias-aligned samples in the training set ($\rho$)}} \\
\cmidrule(lr){2-7}
& \multicolumn{2}{c}{$\rho = 0.99$}
& \multicolumn{2}{c}{$\rho = 0.995$}
& \multicolumn{2}{c}{$\rho = 0.997$} \\
\cmidrule(lr){2-7}
& \sp\ (\%) $\uparrow$ & MFLOPs $\downarrow$
& \sp\ (\%) $\uparrow$ & MFLOPs $\downarrow$
& \sp\ (\%) $\uparrow$ & MFLOPs $\downarrow$ \\
\midrule
Vanilla & 0 & 415.4 & 0 & 415.4 & 0 & 415.4 \\
\midrule

BISE
(\cref{tab:biased_mnist_sparsity_flops})
    & $20.9_{\pm 4.4}$ & $328.3_{\pm 18.2}$ 
     & $29.9_{\pm 7.5}$ & $290.8_{\pm 31.0}$ 
     & $35.0_{\pm 1.5}$ & $269.6_{\pm 6.2}$ \\

\makecell[l]{BISE\\[-3pt]\hspace{2pt}+ trainable BN\\[-3pt]\hspace{8pt}(avg., std.)}
& $18.3_{\pm 6.8}$ & $339.2_{\pm 28.1}$ & $26.5_{\pm 4.0}$ & $305.0_{\pm 16.7}$ & $33.3_{\pm 7.9}$ & $276.9_{\pm 32.9}$  \\
    
\makecell[l]{BISE\\[-3pt]\hspace{2pt}+ trainable BN\\[-3pt]\hspace{8pt}(avg., std., $\beta$, $\gamma$)}
& $32.7_{\pm 7.7}$ & $279.5_{\pm 32.0}$ & $45.7_{\pm 3.9}$ & $225.6_{\pm 16.2}$ & $48.2_{\pm 12.0}$ & $214.9_{\pm 49.8}$  \\

\bottomrule
\end{tabular}
}
\end{table}



\section{The auxiliary parameters \texorpdfstring{$\{\boldsymbol{m_i}\}$}{\{m\_i\}} provide a way to rank neurons}

\begin{figure}[h]
\includegraphics[width=\columnwidth]{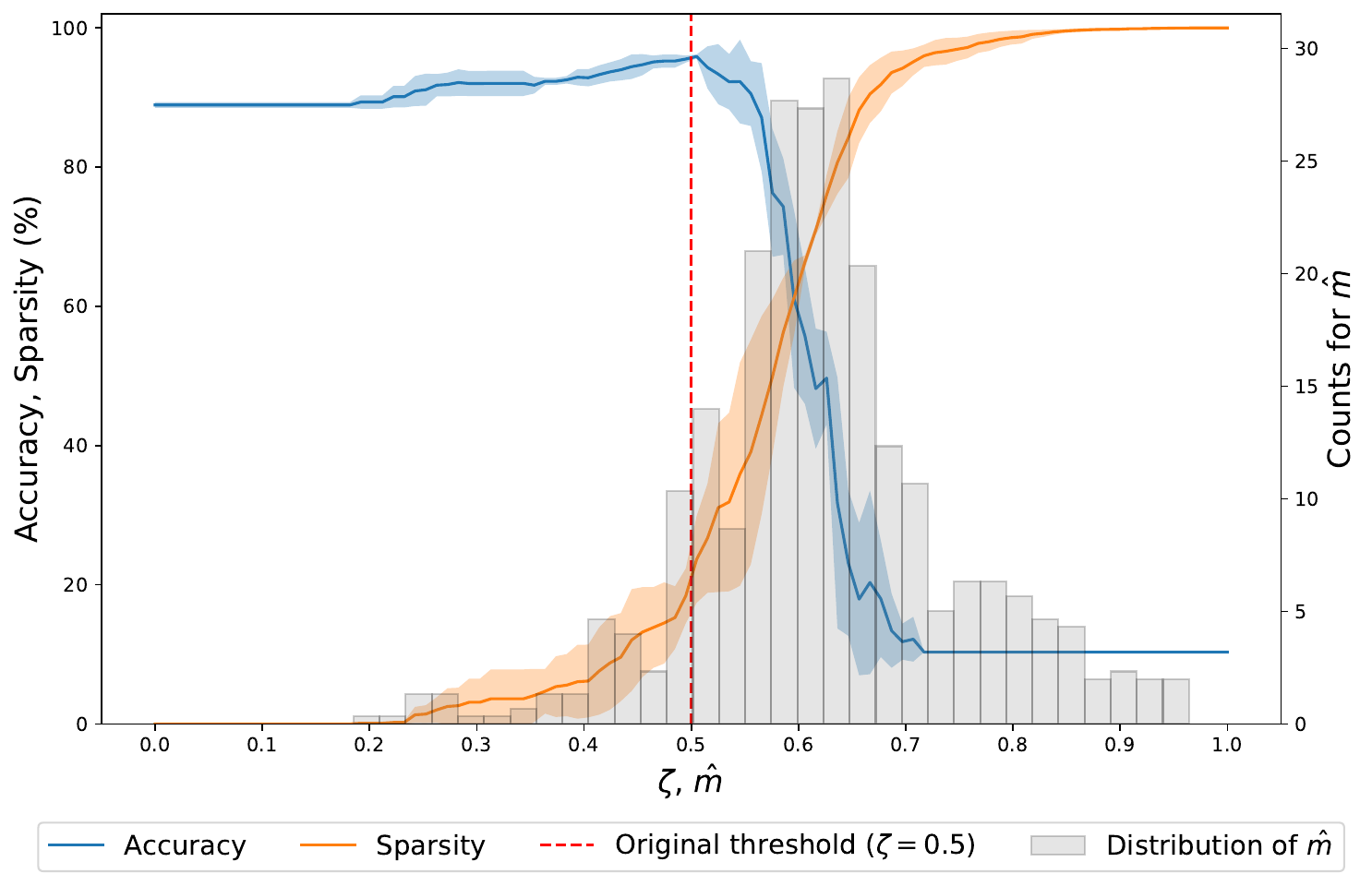}
\centering
\caption{Analysis of the effect of varying the threshold $\zeta$ on the pruned network. Threshold $\zeta=0$ corresponds to the vanilla dense model; $\zeta=0.5$ is the original threshold used in BISE.
}
\label{fig_rank_pruning_all}
\end{figure}

In this section, we show that the auxiliary variables $\{m_i\}$ represent a \textit{ranking} of neurons/filters, indicating the priority with which we should prune them to accomplish the debiasing.
Given a trained set $\{m_i\}$, let us modify the gating function
as $\hat{h}_i = h_i \cdot \mathbf{1}\{\hat{m}_i \geq \zeta\}$, with $~{\zeta\in[0,1]}$ being the threshold for defining the boolean pruning mask, and considering $\hat{m}_i$ at $\tau=1$, \textit{i.e.}, $~{\hat{m}_i = \sigma(m_i)}$.
For the BiasedMNIST dataset with $\rho=0.99$,
Fig.~\ref{fig_rank_pruning_all} shows the distribution of $\hat{m}_i$ in the debiased model, as well as how the sparsity and the performance on the unbiased test set vary as we modify $\zeta$. In particular, no neurons are pruned for $~{\zeta=0}$, while all the neurons are pruned for $~{\zeta=1}$. As we observe, for $\zeta=0$, the accuracy corresponds to the performance of the vanilla model,
and, as we increase $\zeta$, we achieve a maximum performance
around $\zeta=0.5$ (which consists of the threshold originally used during the training of $\{m_i\}$). In the sequence, as $\zeta$ approaches 1, the accuracy progressively drops to random guess (10\%), as expected, since we perform pruning without finetuning the remaining network parameters. 
We observe that the steepest drop in accuracy and increase in sparsity happen when $\zeta$ is around the mode of the distribution of $\hat{m}_i$, due to the removal of a large number of parameters.

\section{Unsupervised debiasing on BiasedMNIST}
\label{sec:unsupervised_bise}

In \cref{tab:table_unsupervised_bmnist} and \cref{tab:table_unsupervised_bmnist_sparsity_flops}, we present the results for BISE on BiasedMNIST, in the \textit{unsupervised} debiasing scenario.
The identification model, $\tilde f$, is trained for only one epoch. The predictions $\tilde b$ from $\tilde f$ closely reflect the true color $b$, hence making the unsupervised procedure close to the supervised debiasing case. Our results are generally competitive with other unsupervised debiasing approaches, such as LfF~\cite{nam2020learning} and SoftCon~\cite{biascon} (\cf 
\cref{tab:table_biased_mnist}
in the main paper).


\begin{table}[!h]
\centering
\caption{Unsupervised (\textit{Unsup.}) debiasing on BiasedMNIST.}
\label{tab:table_unsupervised_bmnist}
\resizebox{0.8\columnwidth}{!}{
\begin{tabular}{lccc}
\toprule
\multirow{2}{*}{\textbf{Method}}          & \multicolumn{3}{c}{\textbf{Accuracy (\%)}} \\
                                 \cmidrule{2-4}
                                 & $\rho=0.99$ & $\rho=0.995$ & $\rho=0.997$ \\
\midrule
Vanilla                          & $88.9_{\pm 0.4}$         & $75.1_{\pm 4.2}$         & $66.1_{\pm 1.7}$      \\
\midrule

\textit{Unsup.} BISE  & $95.8_{\pm 0.5}$          & $92.0_{\pm 2.0}$         & $90.4_{\pm 0.6}$      \\

\textit{Unsup.} BISE (last)  & $95.4_{\pm 0.7}$      & $90.6_{\pm 1.9}$     & $90.0_{\pm 0.2}$          \\

\bottomrule
\end{tabular}
}%
\end{table}


\begin{table}[!h]
\centering
\caption{Sparsity (\sp) and computational cost (MFLOPs) of models, for the BiasedMNIST dataset, in the unsupervised debiasing scenario.}
\label{tab:table_unsupervised_bmnist_sparsity_flops}
\resizebox{\columnwidth}{!}{
\begin{tabular}{
    l
    @{\hskip 8pt}
    cc
    @{\hskip 15pt}
    cc
    @{\hskip 15pt}
    cc
}
\toprule
\multirow{3}{*}{\textbf{Method}} 
& \multicolumn{6}{c}{\textbf{Proportion of bias-aligned samples in the training set ($\rho$)}} \\
\cmidrule(lr){2-7}
& \multicolumn{2}{c}{$\rho = 0.99$}
& \multicolumn{2}{c}{$\rho = 0.995$}
& \multicolumn{2}{c}{$\rho = 0.997$} \\
\cmidrule(lr){2-7}
& \sp\ (\%) $\uparrow$ & MFLOPs $\downarrow$
& \sp\ (\%) $\uparrow$ & MFLOPs $\downarrow$
& \sp\ (\%) $\uparrow$ & MFLOPs $\downarrow$ \\
\midrule
Vanilla & 0 & 415.4 & 0 & 415.4 & 0 & 415.4 \\
\midrule

\textit{Unsup.} BISE & $19.9_{\pm 4.4}$ & $332.4_{\pm 18.2}$ & $26.2_{\pm 4.7}$ & $306.3_{\pm 19.5}$ & $30.9_{\pm 2.3}$ & $286.6_{\pm 9.7}$  \\
    
\textit{Unsup.} BISE (last) & $19.7_{\pm 3.6}$ & $333.4_{\pm 14.9}$ & $28.8_{\pm 4.9}$ & $295.6_{\pm 20.3}$ & $28.8_{\pm 4.5}$ & $295.5_{\pm 18.7}$  \\

\bottomrule
\end{tabular}
}
\end{table}

\section{Robustness to noisy labels}

\begin{figure}[h]
\centering

\includegraphics[width=0.9\columnwidth]{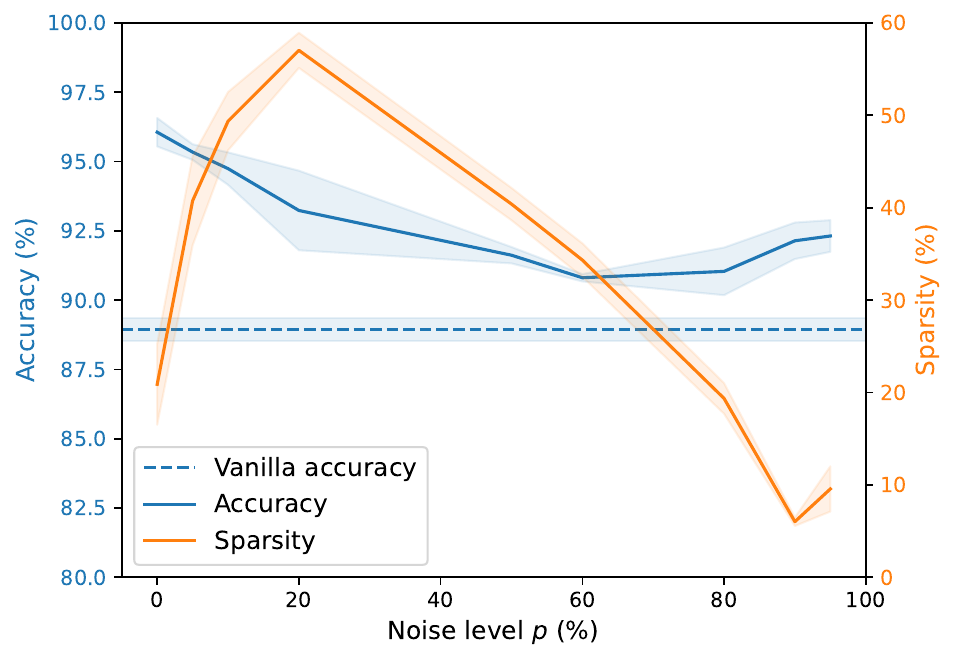}

\caption{Results on BiasedMNIST (training set with $\rho=0.99$) under noisy bias labels. Noise level $p=0\%$ corresponds to the standard setting (\textit{i.e.}, use of true bias labels).}
\label{fig:fig_noisy_exps}
\end{figure}

To complement the study from Appendix~\ref{sec:unsupervised_bise},
we have conducted experiments on BiasedMNIST, with $\rho=0.99$, to show how BISE is robust to \textit{noisy} bias labels -- which can be usual in practical applications.
We proceed as follows: before applying BISE, we randomly select a fraction $p\in[0,1]$ of the training samples, and change their bias label $b$ to another random value.
The results for test accuracy are reported in \cref{fig:fig_noisy_exps}. We observe that, for all noise levels considered, BISE could improve the test accuracy, in comparison to the vanilla model, even if by a small margin.
Importantly, even under high noise, our method does not extract a subnetwork that is worse than the vanilla model.

\section{On the pruning of bias-related features}

As a simple empirical assessment of whether the pruned parameters correspond to bias-relevant features, we have conducted the following experiment. We consider the BiasedMNIST setup and the two following models:
\begin{itemize}
    \item a vanilla model, trained on a biased set with $\rho=0.99$;
    \item the corresponding debiased subnetwork, obtained with BISE.
\end{itemize}
For each model, we attach an auxiliary classifier to the output of the corresponding encoder (\textit{i.e.}, right before the last layer); the auxiliary classifiers are trained to predict the color of the digits, $b$, based on the latent representation from the respective encoder’s output.
In this specific study, we perform this training using an \textit{unbiased} version of the BiasedMNIST training set (\textit{i.e.}, with $\rho=0.1$), to avoid the auxiliary classifiers from leveraging digit-related features (due to the correlation digit-color in a biased set). The training is performed for 100 epochs, using SGD with learning rate 0.1, weight decay $10^{-4}$, momentum 0.9. The results on color prediction by the auxiliary classifiers are reported in \cref{tab:aux_classif_study}.

\begin{table}[h]
\centering
\caption{Comparison of accuracy on color prediction for two auxiliary classifiers: one trained on top of a vanilla encoder, and the other trained on top of the BISE-pruned encoder.}
\label{tab:aux_classif_study}
\resizebox{0.85\columnwidth}{!}{
\begin{tabular}{l c}
\toprule
\textbf{Model} 
& \makecell[c]{\textbf{Accuracy of the aux. classifier}\\[-2pt]\textbf{on color prediction (\%)}} \\
\midrule
Vanilla & $97.6_{\pm 1.0}$ \\
Debiased with BISE & $82.1_{\pm 7.3}$ \\
\bottomrule
\end{tabular}
} %
\end{table}

We observe that, although the bias-related features (\textit{i.e.}, related to the color) are not entirely removed from the latent representation provided by the BISE-pruned encoder, in this case, the color prediction is made ``harder'', as indicated by the significantly lower color prediction accuracy. This observation suggests that, by applying our method to prune the network, we can reduce the impact of bias on the information that can be leveraged by a classifier attached to the encoder. In other words, it is suggested that, by applying BISE, the representation $\hat z$ provided by the pruned encoder becomes more invariant to the bias, hence making it harder to predict $b$ from $\hat z$.

\section{Summary table}

In \cref{tab:summary_table}, we present a summary of the results across the different datasets considered. The competing methods correspond to the ones presented in the tables from 
\cref{sec:experiments:results}.

\begin{table*}[t]
\centering
\footnotesize
\setlength{\tabcolsep}{3.5pt}
\resizebox{\textwidth}{!}{%
\begin{tabular}{llccccc}
\toprule
\multirow{2}{*}{\textbf{Metric}} & \multirow{2}{*}{\textbf{Model}} & \multicolumn{5}{c}{\textbf{Dataset}} \\
\cmidrule(lr){3-7}
 &  & \makecell{\textbf{BiasedMNIST}\\($\rho=0.99$)}
 & \makecell{\textbf{Corrupted-CIFAR10}\\($\rho=0.95$)}
 & \textbf{CelebA}
 & \makecell{\textbf{Multi-Color MNIST}\\($\rho_\text{L}=0.99$, $\rho_\text{R}=0.95$)}
 & \textbf{CivilComments} \\
\midrule

\multirow{4}{*}{\textbf{Acc. (\%)}} 
& Vanilla & $88.9_{\pm 0.4}$ & $47.18_{\pm 0.34}$ & $76.5_{\pm 2.1}$ & $58.2_{\pm 0.6}$ & $59.6_{\pm 2.7}$ \\
& Best competitor & $98.1$ (BCon+BBal~\cite{biascon}) & $51.13$ (DFA~\cite{dataaug1}) & $91.4$ (BCon+BBal~\cite{biascon}) & $73.1$ (VCBA~\cite{nahon2023mining}) & $80.4$ (Group DRO~\cite{sagawa2019distributionally}) \\
& BISE$^\ast$ & $96.1_{\pm 0.5}$ & $55.38_{\pm 1.96}$ & $89.7_{\pm 0.8}$ & $60.3_{\pm 1.0}$ & $80.4_{\pm 0.2}$ \\
& BISE + finetuning$^\ast$ & $98.1_{\pm 0.1}$ & $57.22_{\pm 1.81}$ & $91.8_{\pm 1.3}$ & $70.6_{\pm 1.6}$ & $81.0_{\pm 0.1}$ \\
\midrule

\textbf{\sp\ (\%)}
& BISE$^\ast$ & $20.9_{\pm 4.4}$ & $82.3_{\pm 0.6}$ & $67.6_{\pm 0.8}$ & $17.1_{\pm 5.3}$ & $26.0_{\pm 5.4}$ \\
\midrule

\multirow{2}{*}{\textbf{FLOPs}}
& Vanilla/competitors & $415.4$ M & $37.1$ M & $1818.6$ M & $256.2$ k & -- \\
& BISE$^\ast$ & $328.3_{\pm 18.2}$ M & $22.5_{\pm 0.6}$ M & $821.5_{\pm 33.1}$ M & $212.3_{\pm 13.5}$ k & -- \\
\bottomrule
\end{tabular}
}
\caption{Summary of results across datasets.
($^\ast$) indicates that the result for \textit{BISE ``best''} is reported whenever a validation set is available; otherwise, we report the result for \textit{BISE ``last''}.
The metric \textit{Acc.} corresponds to the accuracy on an unbiased test set, in the case of BiasedMNIST, CelebA and Corrupted-CIFAR10; for Multi-Color MNIST, it denotes the average test accuracy across the four existing groups in the dataset
(as described in \cref{sec:experiments:results});
for CivilComments, it denotes the worst-group test accuracy. The sparsity (\sp) for the vanilla and competing methods is zero, since no parameter is removed from the network.
}
\label{tab:summary_table}
\end{table*}

\clearpage

\end{document}